\documentclass{article}

\usepackage{arxiv}

\usepackage[utf8]{inputenc} 
\usepackage[T1]{fontenc}    
\usepackage{hyperref}       
\usepackage{url}            
\usepackage{booktabs}       
\usepackage{amsfonts}       
\usepackage{nicefrac}       
\usepackage{microtype}      
\usepackage{graphicx}
\usepackage{natbib}
\usepackage{doi}
\usepackage{subcaption}
\usepackage{amssymb, amsmath, amsthm}
\usepackage{algorithm}
\usepackage{algpseudocode}
\usepackage{appendix}
\usepackage{enumitem}
\usepackage{graphbox}
\title{Sparse Factorization of Large Square Matrices}

\date{} 					

\newcommand*\samethanks[1][\value{footnote}]{\footnotemark[#1]}
\author{ Ruslan Khalitov\thanks{Equal contribution}, \  Tong Yu\samethanks, \  Lei Cheng, \ Zhirong Yang\thanks{Corresponding author, \texttt{zhirong.yang@ntnu.no}}\\
Norwegian University of Science and Technology}

\renewcommand{\shorttitle}{Sparse Factorization of Large Square Matrices}

\hypersetup{
	pdftitle={Sparse Factorization of Large Square Matrices},
	pdfsubject={},
	pdfauthor={Ruslan Khalitov, Tong Yu, Lei Cheng, and Zhirong Yang},
	pdfkeywords={matrix factorization, square matrix, attention, sparse},
}

\begin{document}
	\maketitle

\newcommand{\matA}{\mathbf{A}}
\newcommand{\matB}{\mathbf{B}}
\newcommand{\matC}{\mathbf{C}}
\newcommand{\matD}{\mathbf{D}}
\newcommand{\matE}{\mathbf{E}}
\newcommand{\matF}{\mathbf{F}}
\newcommand{\matG}{\mathbf{G}}
\newcommand{\matH}{\mathbf{H}}
\newcommand{\matI}{\mathbf{I}}
\newcommand{\matK}{\mathbf{K}}
\newcommand{\matL}{\mathbf{L}}
\newcommand{\matM}{\mathbf{M}}
\newcommand{\matN}{\mathbf{N}}
\newcommand{\matO}{\mathbf{O}}
\newcommand{\matP}{\mathbf{P}}
\newcommand{\matQ}{\mathbf{Q}}
\newcommand{\matR}{\mathbf{R}}
\newcommand{\matS}{\mathbf{S}}
\newcommand{\matT}{\mathbf{T}}
\newcommand{\matU}{\mathbf{U}}
\newcommand{\matV}{\mathbf{V}}
\newcommand{\matW}{\mathbf{W}}
\newcommand{\matX}{\mathbf{X}}
\newcommand{\matY}{\mathbf{Y}}
\newcommand{\matZ}{\mathbf{Z}}
\newcommand{\matg}{\mathbf{g}}

\newcommand{\calA}{\mathcal{A}}
\newcommand{\calB}{\mathcal{B}}
\newcommand{\calC}{\mathcal{C}}
\newcommand{\calD}{\mathcal{D}}
\newcommand{\calE}{\mathcal{E}}
\newcommand{\calF}{\mathcal{F}}
\newcommand{\calG}{\mathcal{G}}
\newcommand{\calH}{\mathcal{H}}
\newcommand{\calI}{\mathcal{I}}
\newcommand{\calJ}{\mathcal{J}}
\newcommand{\calK}{\mathcal{K}}
\newcommand{\calL}{\mathcal{L}}
\newcommand{\calM}{\mathcal{M}}
\newcommand{\calN}{\mathcal{N}}
\newcommand{\calO}{\mathcal{O}}
\newcommand{\calP}{\mathcal{P}}
\newcommand{\calQ}{\mathcal{Q}}
\newcommand{\calR}{\mathcal{R}}
\newcommand{\calS}{\mathcal{S}}
\newcommand{\calT}{\mathcal{T}}
\newcommand{\calU}{\mathcal{U}}
\newcommand{\calV}{\mathcal{V}}
\newcommand{\calW}{\mathcal{W}}
\newcommand{\calX}{\mathcal{X}}
\newcommand{\calY}{\mathcal{Y}}
\newcommand{\calZ}{\mathcal{Z}}

\newcommand{\bbA}{\mathbb{A}}
\newcommand{\bbB}{\mathbb{B}}
\newcommand{\bbR}{\mathbb{R}}
\newcommand{\bbZ}{\mathbb{Z}}
\newcommand{\bbE}{\mathbb{E}}
\newcommand{\bbH}{\mathbb{H}}

\newcommand{\veca}{\mathbf{a}}
\newcommand{\vecb}{\mathbf{b}}
\newcommand{\vecc}{\mathbf{c}}
\newcommand{\vecd}{\mathbf{d}}
\newcommand{\vece}{\mathbf{e}}
\newcommand{\vecf}{\mathbf{f}}
\newcommand{\vecg}{\mathbf{g}}
\newcommand{\vech}{\mathbf{h}}
\newcommand{\veci}{\mathbf{i}}
\newcommand{\vecj}{\mathbf{j}}
\newcommand{\veck}{\mathbf{k}}
\newcommand{\vecl}{\mathbf{l}}
\newcommand{\vecm}{\mathbf{m}}
\newcommand{\vecn}{\mathbf{n}}
\newcommand{\veco}{\mathbf{o}}
\newcommand{\vecp}{\mathbf{p}}
\newcommand{\vecq}{\mathbf{q}}
\newcommand{\vecr}{\mathbf{r}}
\newcommand{\vecs}{\mathbf{s}}
\newcommand{\vect}{\mathbf{t}}
\newcommand{\vecu}{\mathbf{u}}
\newcommand{\vecv}{\mathbf{v}}
\newcommand{\vecw}{\mathbf{w}}
\newcommand{\vecx}{\mathbf{x}}
\newcommand{\vecy}{\mathbf{y}}
\newcommand{\vecz}{\mathbf{z}}

\newcommand{\vecalpha}{\boldsymbol{\alpha}}
\newcommand{\vecbeta}{\boldsymbol{\beta}}
\newcommand{\veceta}{\boldsymbol{\eta}}
\newcommand{\vectheta}{\boldsymbol{\theta}}
\newcommand{\vecphi}{\boldsymbol{\phi}}
\newcommand{\vecpsi}{\boldsymbol{\psi}}
\newcommand{\vecrho}{\boldsymbol{\rho}}
\newcommand{\vectau}{\boldsymbol{\tau}}
\newcommand{\vecmu}{\boldsymbol{\mu}}
\newcommand{\veceps}{\boldsymbol{\epsilon}}
\newcommand{\vecxi}{\boldsymbol{\xi}}
\newcommand{\vecPhi}{\boldsymbol{\Phi}}
\newcommand{\vecDelta}{\boldsymbol{\Delta}}

\newcommand{\matDelta}{\boldsymbol{\Delta}}
\newcommand{\matEta}{\boldsymbol{\eta}}
\newcommand{\matOmega}{\boldsymbol{\Omega}}
\newcommand{\matPhi}{\boldsymbol{\Phi}}
\newcommand{\matPsi}{\boldsymbol{\Psi}}
\newcommand{\matTheta}{\boldsymbol{\Theta}}
\newcommand{\matLambda}{\boldsymbol{\Lambda}}
\newcommand{\matSigma}{\boldsymbol{\Sigma}}
\newcommand{\matzero}{\mathbf{0}}
\newcommand{\IndexSetI}{\mathcal{I}}
\newcommand{\grad}{\mathcal{\nabla}}

\newcommand{\vecone}{\mathbf{1}}
\newcommand{\veczero}{\mathbf{0}}

\def\maximize{\mathop{{\mathgroup\symoperators maximize}}}
\def\Maximize{\mathop{{\mathgroup\symoperators Maximize}}}
\def\minimize{\mathop{{\mathgroup\symoperators minimize}}}

\def\approach{\mathop{{\mathgroup\symoperators \longrightarrow}}}
\def\defineoperator{\mathop{{\mathgroup\symoperators =}}}
\newcommand{\define}{\defineoperator^{\text{def}}}

\newcommand{\Tr}{\text{Tr}}
\newcommand{\trace}{\text{trace}}
\newcommand{\diag}{\text{diag}}
\newcommand{\gradWJ}{\nabla_{\scriptscriptstyle{\matW}}\calJ}
\newcommand{\const}{\text{constant}}
\newcommand{\fracpartial}[2]{\frac{\partial #1}{\partial  #2}}

\newcommand{\defeq}{\stackrel{\text{def}}{=}}

\newcommand{\Xh}{\widehat{X}}

\begin{abstract}
Square matrices appear in many machine learning problems and models. Optimization over a large square matrix is expensive in memory and in time. Therefore an economic approximation is needed. Conventional approximation approaches factorize the square matrix into a number matrices of much lower ranks. However, the low-rank constraint is a performance bottleneck if the approximated matrix is intrinsically high-rank or close to full rank. In this paper, we propose to approximate a large square matrix with a product of sparse full-rank matrices. In the approximation, our method needs only $N(\log N)^2$ non-zero numbers for an $N\times N$ full matrix. We present both non-parametric and parametric ways to find the factorization. In the former, we learn the factorizing matrices directly, and in the latter, we train neural networks to map input data to the non-zero matrix entries. The sparse factorization method is tested for a variety of synthetic and real-world square matrices. The experimental results demonstrate that our method gives a better approximation when the approximated matrix is sparse and high-rank. Based on this finding, we use our parametric method as a scalable attention architecture that performs strongly in learning tasks for long sequential data and defeats Transformer and its several variants. Our code is publicly available\footnote{https://github.com/RuslanKhalitov/SparseFactorization}.
\end{abstract}


\section{Introduction}
Many machine learning models include one or more square matrices, such as the kernel matrix in Support Vector Machines \citep{svm} or Gaussian Process, the affinity matrix in graph or network representation, and the attention matrix in transformers \citep{transformer}. In many machine learning problems, the solution space is also over square matrices, for example, graph matching and network architecture optimization.

Obtaining a full $N\times N$ matrix can cause infeasible storage and computational cost if $N$ is large. For example, a double-precision kernel matrix of the typical \texttt{MNIST} handwritten digit data set ($N=70000$) requires about 36.5G memory. In another example, we need to calculate eight quintillion float numbers to fill a full attention matrix of a human DNA sequence with about 3.2 billion base pairs.

Therefore we need economical surrogates to approximate the full square matrices at a large scale. Matrix factorization is a widely used approach that approximately factorizes the square matrix to some cheaper matrices. In conventional matrix factorization, the factorizing matrices must be low-rank, for example, in Truncated Singular Value Decomposition and Nystr\"o{}m approximation \citep{nystrom1,nystrom2}. However, these conventional approaches do not work well if the square matrix in the approximation is not low-rank.

This paper proposes a novel approximation method called Sparse Factorization (SF), where the approximated square matrix is factorized into some full-rank sparse matrices. Using the Chord protocol \citep{chord} to specify the non-zero entry positions\footnote{Non-zero entries are those stored, including both non-zeros and explicit zeros.}, we can match an $N\times N$ full matrix with $\log N$ factorizing matrices, where each contains $N\log N$ non-zero entries and thus in total $N(\log N)^2$ non-zeros. Therefore the approximation is economical when $N$ is large.

We first study non-parametric learning of the approximation. That is, we directly minimize the approximation loss over the sparse factorizing matrices. The new method was tested on a variety of synthetic and real-world square matrices, with comparison to the most accurate low-rank approximation method, TSVD. We find that SF is especially suitable for approximating sparse square matrices with unknown non-zero positions, with a clear win over TSVD.

The Chord protocol also enables us to parameterize the mapping from input data to the non-zero entries in a matrix row with neural networks. This parametric form of SF thus provides a new attention architecture. We find that only one block of SF attention defeats Transformer and its several state-of-the-art variants on very long synthetic sequence classification and on the public Long Range Arena benchmark tasks.

The remaining of the paper is organized as follows. We first briefly review the matrix approximation based on low-rank factorization in Section \ref{sec:lowrank}. In Section \ref{sec:nonparametric}, we present the machine learning formulation of sparse factorization of square matrices. The parametric formulation using neural networks is proposed in Section \ref{sec:parametric}. In Section \ref{sec:exps}, we present the experimental settings and results. Conclusion and future work are given in Section \ref{sec:conclusion}.

\section{Low-Rank Matrix Factorization}
\label{sec:lowrank}
The conventional matrix approximation is to factorize the large square matrix $X$ into a few low-rank matrices, for example $X\approx \Xh = WH$ or $X\approx WSH$ with $W\in\bbR^{N\times r}$, $S\in\bbR^{r\times r}$, $H\in\bbR^{r\times N}$ and $r\ll N$. Nystr\"{o}m decomposition \citep{nystrom1} is of the tri-factor kind, where $W$ and $H$ are calculated using normal kernel function, and $S$ is learned.

\begin{figure}[t]
	\begin{center}
		\includegraphics[align=c, width=5cm]{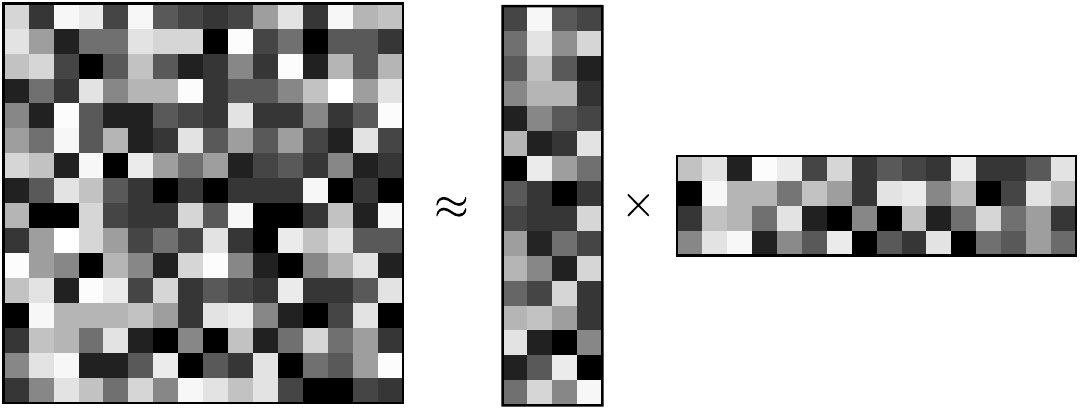}\quad\quad\quad\quad
		\includegraphics[align=c, width=7cm]{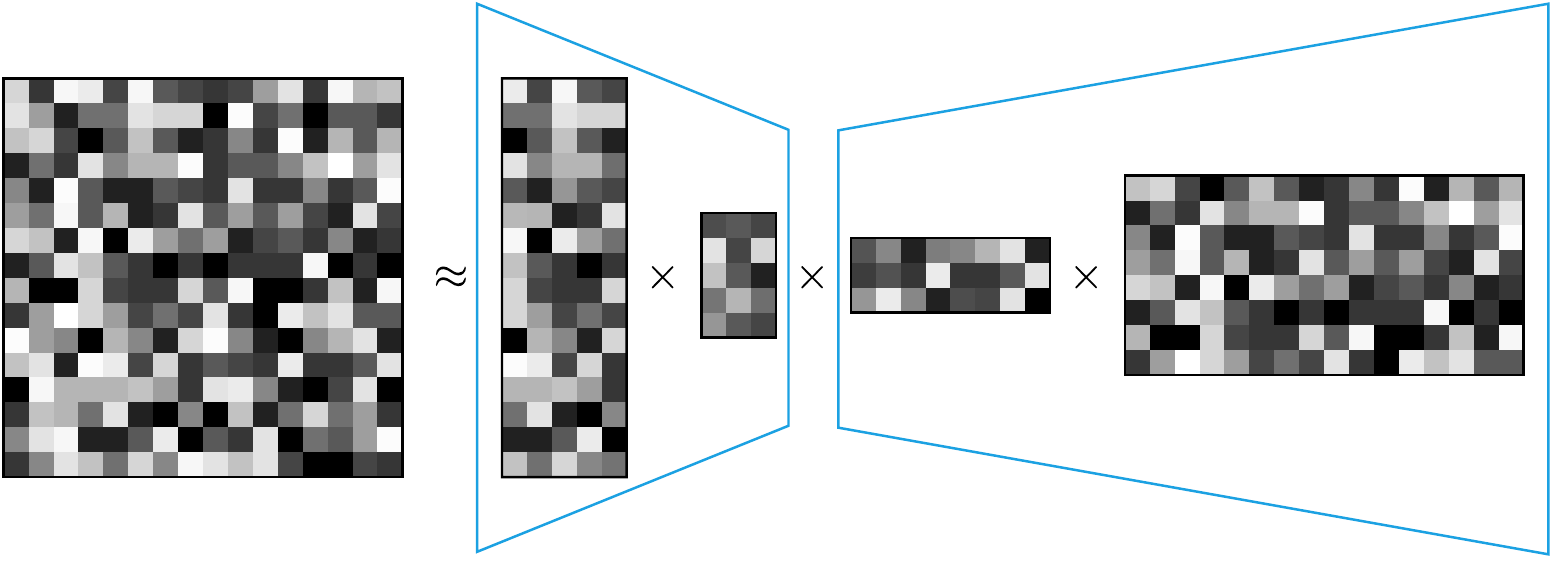}
	\end{center}
	\caption{Examples of low-rank matrix dense matrix factorization. Each small square represents a matrix entry. Left is a two-factor factorization and the right is a four-factor factorization. The trapezoids illustrate the ``bottleneck'', i.e.~grouping the factorizing matrices according the place with the smallest rank.}
	\label{fig:lowrank}
\end{figure}

The approximation error can be measured by a certain divergence, for example the squared Frobenius norm (F-norm): $D(X||\Xh)=\|X-\Xh\|_F^2=\sum_{ij}\left(X_{ij}-\Xh_{ij}\right)^2$.
Minimization of $\|X-WH\|_F^2$ over $W$ and $H$ has a closed-form solution using Truncated Singular Value Decomposition (TSVD). Denote $\Lambda$ the diagonal matrix with the largest $r$ singular values. The corresponding left and right singular vectors as columns form matrices $U$ and $V$, respectively. Then the minimum appears by setting $W = U$ and $H=\Lambda V^T$, where we can move any scaling from $W$ to $H$ or vice versa.

In general $\Xh = \prod_{m=1}^M W^{(m)}$, where $W^{(m)}\in\bbR^{r_m,r_{m+1}}$ with $r_1=r_{M+1}=N$. We can always reduce a multi-factor case to the two-factor form by grouping $L=\prod_{m=1}^{m'-1}W^{(m)}$ and $R=\prod_{m=m'}^{M}W^{(m)}$, where $m'=\arg\min_m\left(\{r_m\}_{m=2}^M\right)$ and $X\approx LR$.
It is required that $r_{m'}\ll N$ when approximating large square matrices. Otherwise, the factorizing matrices are still too large. This two-factor form reveals a bottleneck as illustrated in Figure \ref{fig:lowrank} (bottom), which shows that the approximation quality of all variants of low-rank matrix factorization cannot exceed TSVD($X$, $r_m'$) in terms of Frobenius norm.

\section{Sparse Factorization}
\label{sec:nonparametric}
As we have seen, the performance of LRMF is capped due to the low-rank constraint. Its performance is poor if the approximated square matrix is far from low-rank, i.e., the sum of the few largest eigenvalues does not dominate the matrix trace.

Here we propose a new approximation method that is free of the low-rank constraint. Our method implements the approximation with a number ($M$) of \emph{square} and \emph{sparse} factorizing matrices. The approximation is still economical if the total number of non-zero entries is much smaller than $N^2$.

There are many ways to specify the sparse structure (i.e.~the non-zero positions). A good specification should guarantee that the product of the factorizing matrices, $\Xh$, should be a full matrix. The condition prevents that the approximating matrix contains some always-zero entries. In addition, we consider a secondary requirement that each factorizing matrix has the same sparse structure, which provides better symmetry in the approximation.

\begin{figure}[t]
	\begin{center}
		\includegraphics[width=7cm]{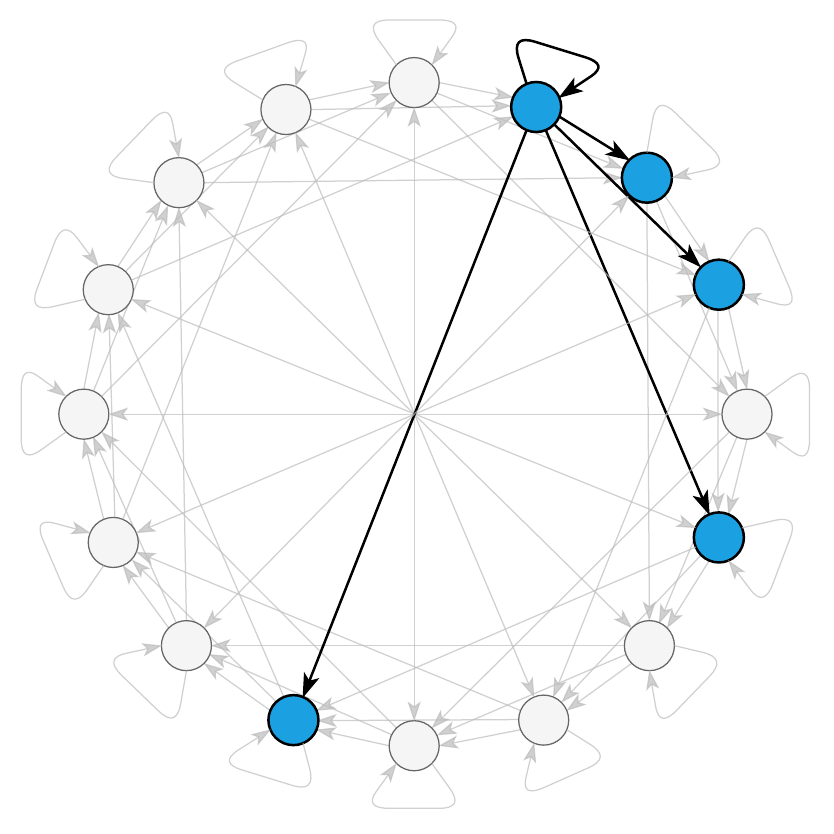}\\[4mm]
		\includegraphics[width=12cm]{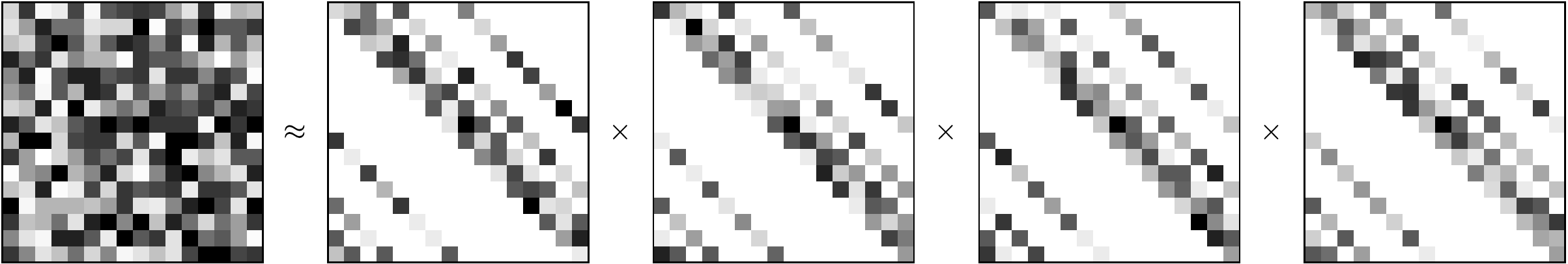}
	\end{center}
	\caption{Illustration of (top) the Chord protocol and (bottom) sparse factorization of a square matrix for $N=16$. Grayscale squares in the factorizing matrices represent stored entries (non-zeros and explicit zeros), and completely white squares represent non-stored entries (implicit zeros).}
	\label{fig:sf16}
	\vspace{-3mm}
\end{figure}

We thus adopt a modified Chord protocol \citep{chord} which originates for peer-to-peer distributed hash tables. In the protocol, the indices from $1$ to $N$ are organized in a circular graph, where the $i$th node connects to itself and the $((i+2^k) \mod N)$-th nodes with $k=0,\dots,K-2$. We set $K=\log_2 N$ in our work. The Chord protocol is illustrated in Figure \ref{fig:sf16} (top).

We use the Chord protocol to specify the non-zero positions in each factorizing matrix, where each matrix row has $\log_2 N$ non-zero entries. Thus every factorizing matrix has $N\log_2 N$ non-zeros. Note that the product of the factorizing matrices corresponds to the connections in the circular graph after multiple hops. We can set the number of factorizing matrices to $M=\log_2 N$, which corresponds to the number of hops, and the resulting matrix product becomes a full matrix with high probability \citep[see][Theorem 2]{chord}. In total, there are $N(\log N)^ 2$ non-zero entries, still much smaller than $N^2$ for a large $N$.

\begin{figure*}[t]
	\begin{center}
		\includegraphics[width=16cm]{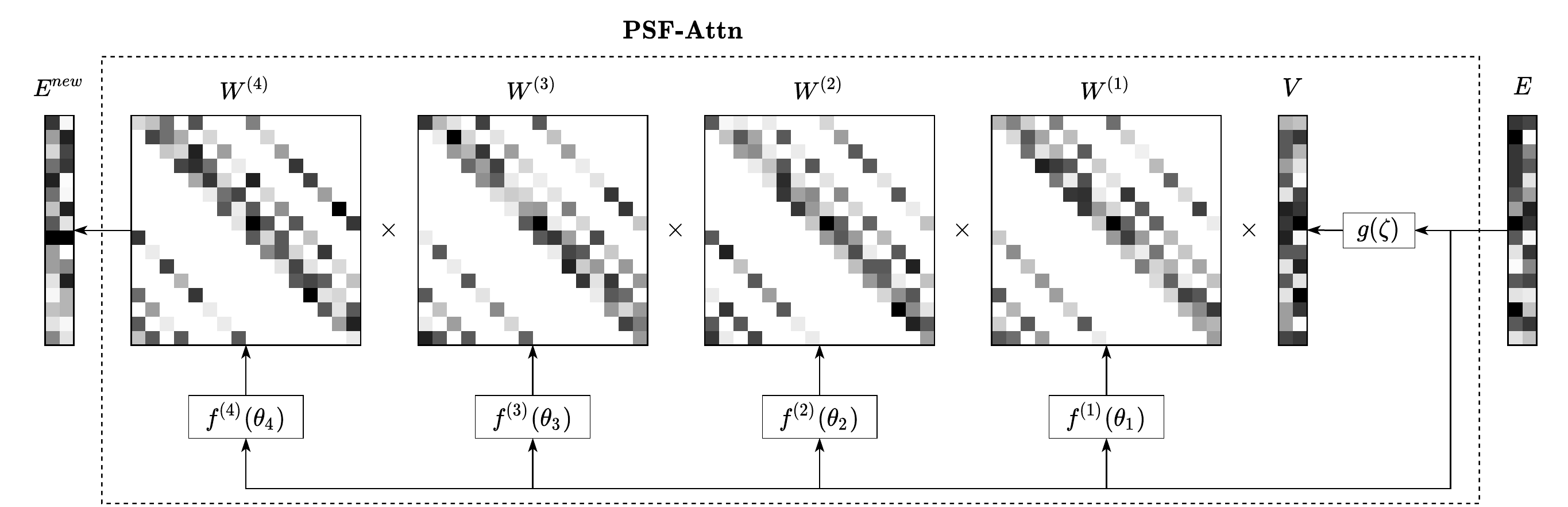}\quad
		\includegraphics[width=4.5cm]{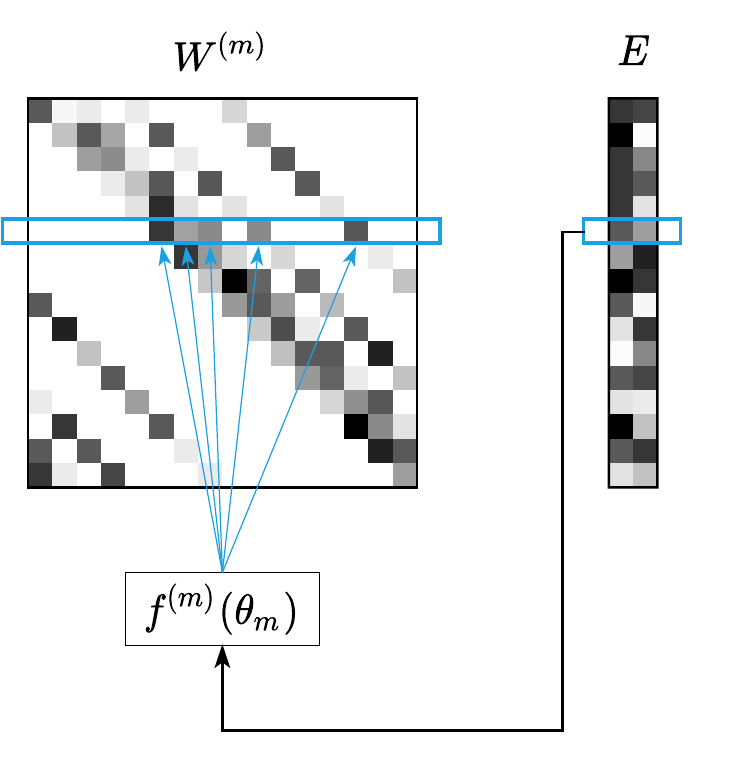}
	\end{center}
	\caption{Illustration of (top) the parametric transformation from current embedding $E$ to new embedding $E^\text{new}$ based on Sparse Factorization and (bottom) setting of MLP output to non-zero entries in the corresponding sparse factorizing matrix.}
	\label{fig:sf16_parametric}
\end{figure*}

The (Chord) Sparse Factorization (SF) can thus be formulated as the following optimization problem:
\begin{align}
\label{eq:np_sf}
    \minimize_{W^{(1)},\dots,W^{(M)}} \left\|X-\prod_{m=1}^M W^{(m)}\right\|_F^2
\end{align}
where $W^{(m)}$'s are sparse square matrices with non-zero positions specified by the Chord protocol. The approximation scheme is illustrated in Figure \ref{fig:sf16} (bottom). We also call the approach non-parametric SF as we directly optimize over the factorizing matrices.

\section{Parametric Sparse Factorization}
\label{sec:parametric}

Besides direct optimization over the factorizing matrices, we can consider the mapping from vectorial input data to the factorizing matrix entries because each node in the Chord protocol has the same out-degree. The mapping as a component can endow the model 1) generalization to newly coming data and 2) representation learning by transforming the current embedding representation to a new embedding.

We present a transformation model as a concrete example to illustrate the idea. It is a transformer-like model (see Figure \ref{fig:sf16_parametric} left), where we replace the scaled dot-product attention with a product of sparse square matrices. The matrix product provides an approximation to a full non-normalized attention matrix.

One block of such Parametric Sparse Factorization Attention (PSF-Attn) transforms data in the current embedding $E$ to new embedding $E^\text{new}=\text{PSF-Attn}(E)$. There is a number of MLPs in the block, where the MLP $f^{(m)}$ with parameters $\theta_m$ takes $E_{i:}$ (the $i$th row of $E$) as an input and returns the non-zeros in the $i$th row of $W^{(m)}$, for $i=1,\dots,N$ and $m=1,\dots,M$. That is, for $k=1,\dots,K$,
\begin{align*}
    W^{(m)}_{ij}=
    \begin{cases}
    \left[f^{(m)}(E_{i:}; \theta_m)\right]_1 & \text{if }j=i \\
    \left[f^{(m)}(E_{i:}; \theta_m)\right]_k & \text{if }j=(i+2^{k-2})\mod N \\
    0 & \text{otherwise}.
    \end{cases}
\end{align*}
Similar to transformers, we use another MLP $g$ with parameters $\zeta$ to convert an $E$ row to the corresponding $V$ row. 

The new model can work as a building block in representation learning frameworks. For example, if we define an objective function $\calJ$ over $E^\text{new}$, the learning can be formulated as the following optimization problem:
\begin{align}
\label{eq:psf}
    \minimize_{\theta_1,\dots,\theta_M,\zeta}~~\calJ(\text{PSF-Attn}(E; \theta_1,\dots,\theta_M,\zeta)),
\end{align}
where the optimization can be implemented with back-propagation and a gradient based algorithm such as Adam \citep{adam}. It is also straightforward to stack multiple PSF-Attn blocks to implement deeper learning.

The PSF-Attn method has two advantages. First, the original transformer and many of its variants  \citep[e.g.][]{linformer,performer,tay2021synthesizer,katharopoulos2020transformers} are built upon scaled dot-product, which essentially employ low-rank matrix factorization. They may not work well if the attention is intrinsically high rank. In contrast, all factorizing matrices in our method are full-rank, and so is their product. Therefore PSF-Attn does not suffer from the low-rank bottleneck constraint. Second, our model does not require softmax over a large square matrix, avoiding many computational difficulties. Removing the softmax also endows more freedom in mixing the $V$ rows because the mixture can go beyond the convex hull.

Our method also differs from the previous multi-layer sparse attention approaches such as \citep{logsparse, child2019generating, correia2019adaptively} because they still use scaled dot-products. PSF-Attn can give full attention in one block. For the $i$th element in the sequence, its attention to other elements can directly be obtained by vector-matrix product $W^{(M)}_{i:}\prod_{m=1}^{M-1}W^{(m)}$.

\section{Experiments}
\label{sec:exps}
We have performed three groups of experiments. In the first group, we tried the non-parametric SF for different types of square matrices and compared it with TSVD. Next, we demonstrated the scalability of PSF-Attn on synthetic sequences up to tens of thousands positions. Finally, we tested PSF-Attn for the Long Range Arena benchmark data sets to see its performance in real-world practice. The first group of experiments was run on a standard PC with an Intel Core i9 CPU. The second and third groups were run on a Linux server with one NVIDIA-Tesla V100 GPU with 32GB of memory.

\subsection{Non-paramatric SF}
\label{sec:exp_nonparametric}

\setlength\tabcolsep{1.5pt}
\newcommand{\sqimgwidth}{2.6cm}
\begin{figure*}[t]
    \centering
    \begin{tabular}{cccccc}
    chess & Lena & apple & skyscraper & lostdoor & ship\\
    \includegraphics[width=\sqimgwidth]{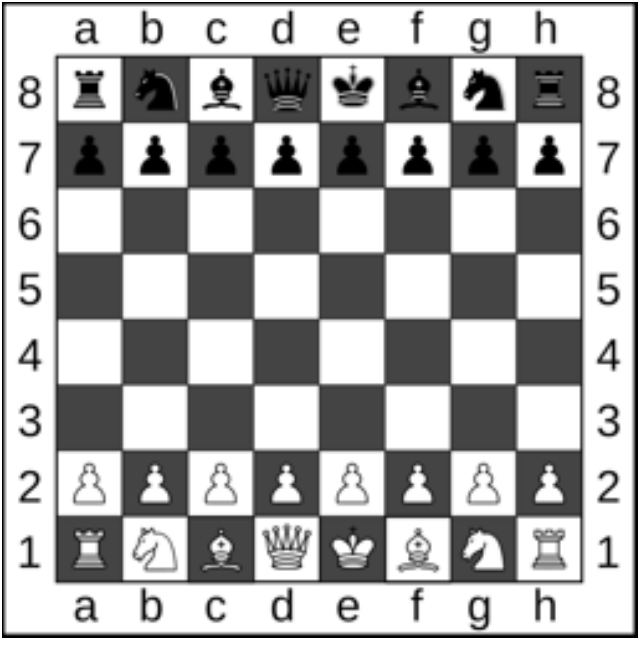} &
    \includegraphics[width=\sqimgwidth]{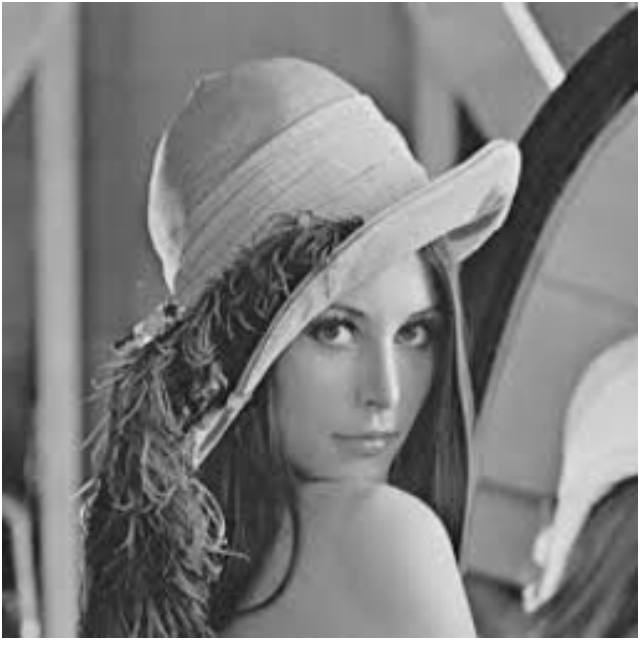} &
    \includegraphics[width=\sqimgwidth]{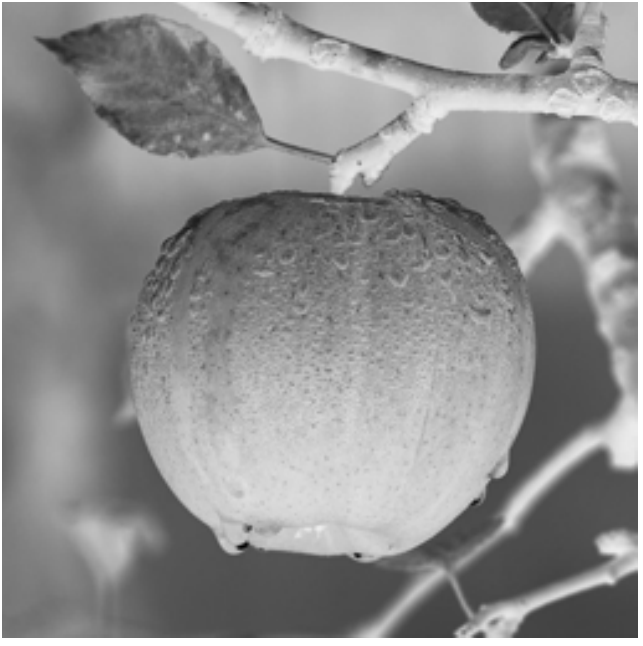} &
    \includegraphics[width=\sqimgwidth]{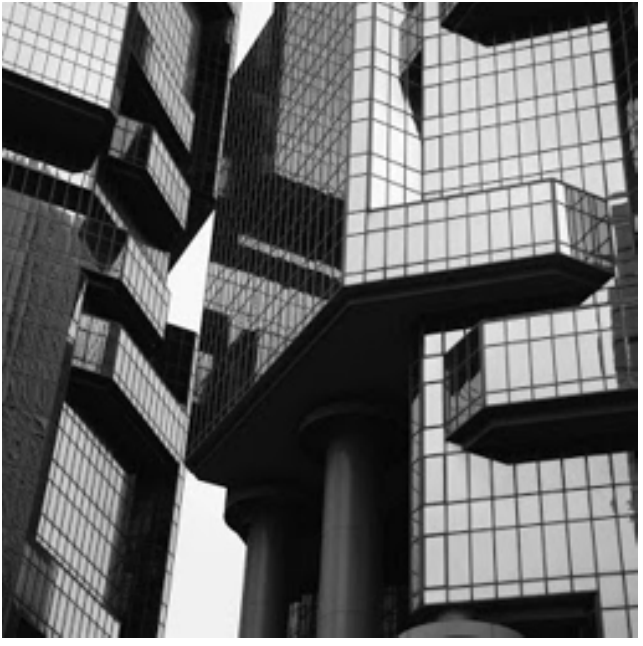} &
    \includegraphics[width=\sqimgwidth]{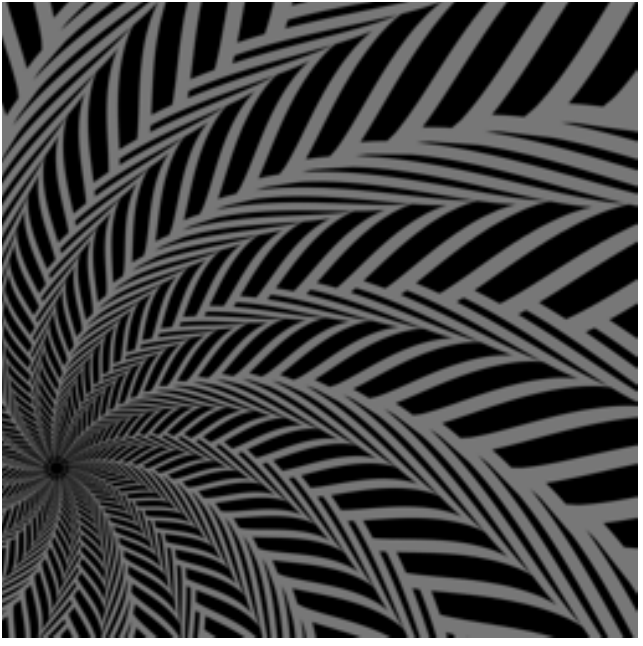} &
    \includegraphics[width=\sqimgwidth]{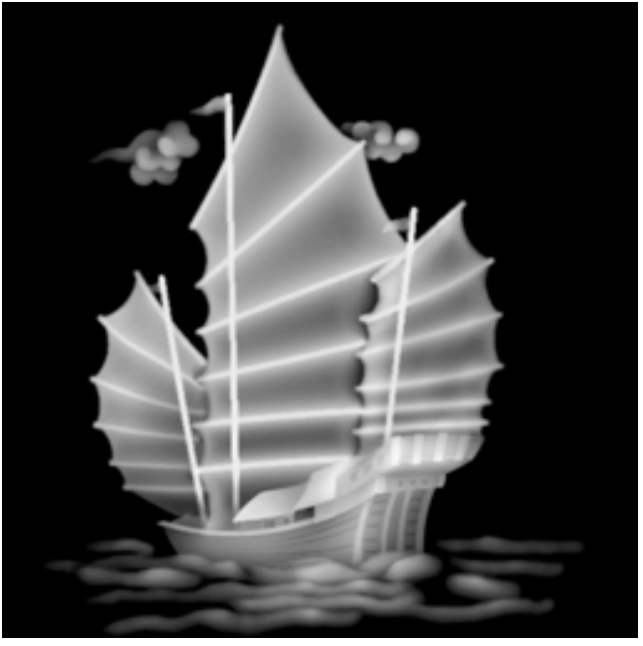} \\[-1.5mm]
    {\tiny {\bf TSVD 126}, SF 136} &
    {\tiny {\bf TSVD 2327}, SF 2649} &
    {\tiny {\bf TSVD 1870}, SF 2297} &
    {\tiny TSVD 4296, {\bf SF 4283}} &
    {\tiny TSVD 6158, {\bf SF 5180}} &
    {\tiny TSVD 1705, {\bf SF 1342}}\\[2mm]
    \includegraphics[width=\sqimgwidth]{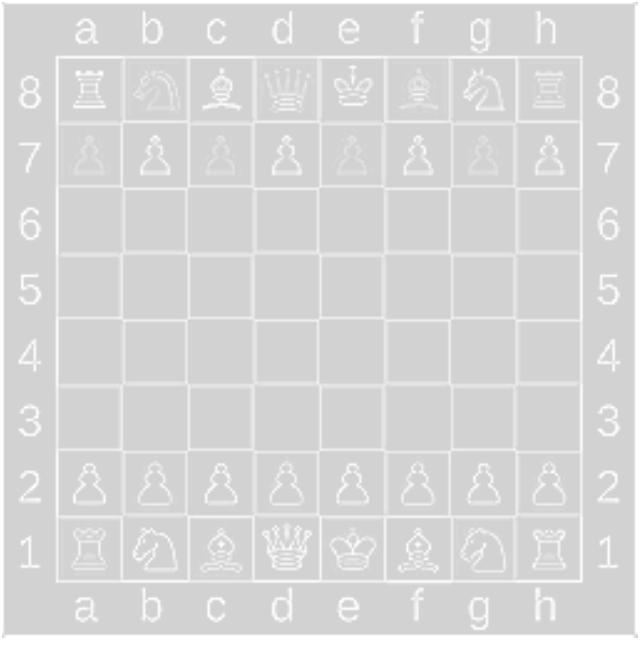} &
    \includegraphics[width=\sqimgwidth]{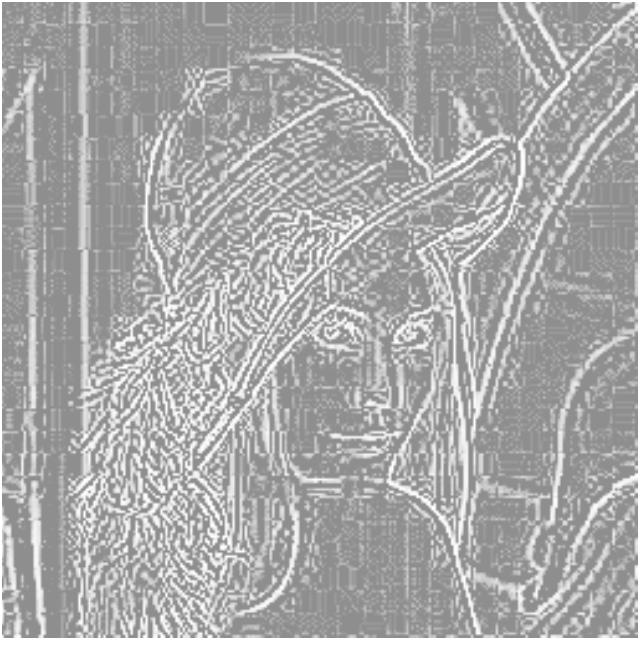} &
    \includegraphics[width=\sqimgwidth]{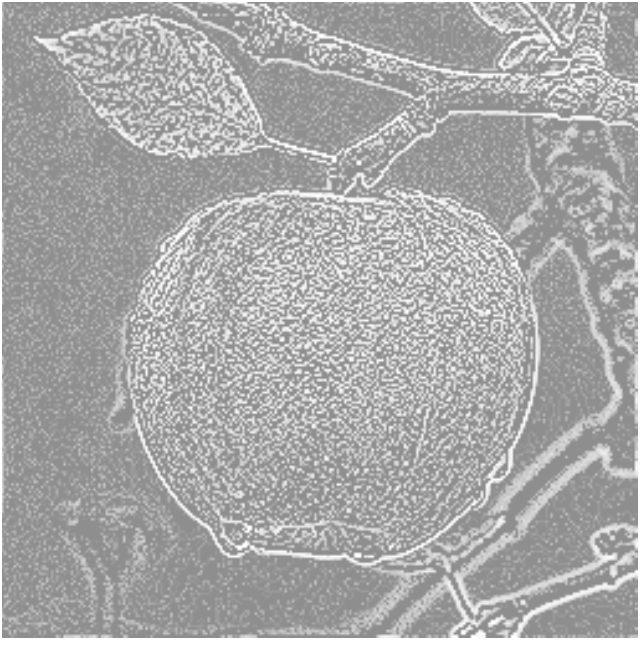} &
    \includegraphics[width=\sqimgwidth]{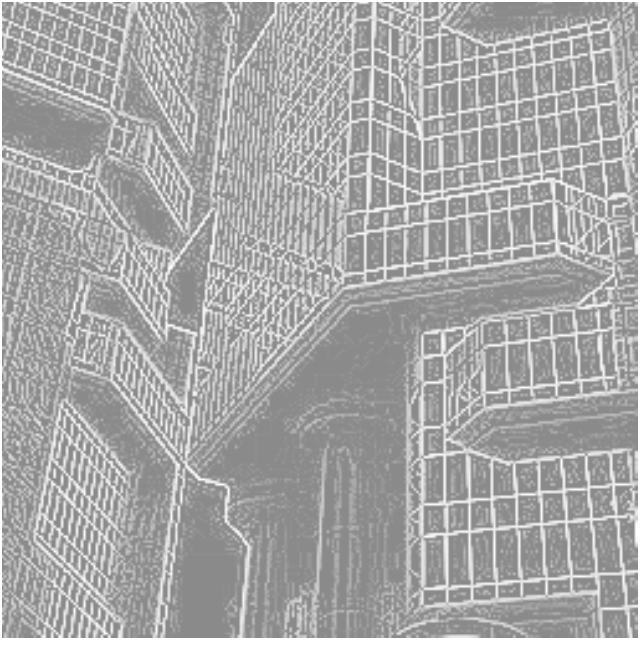} &
    \includegraphics[width=\sqimgwidth]{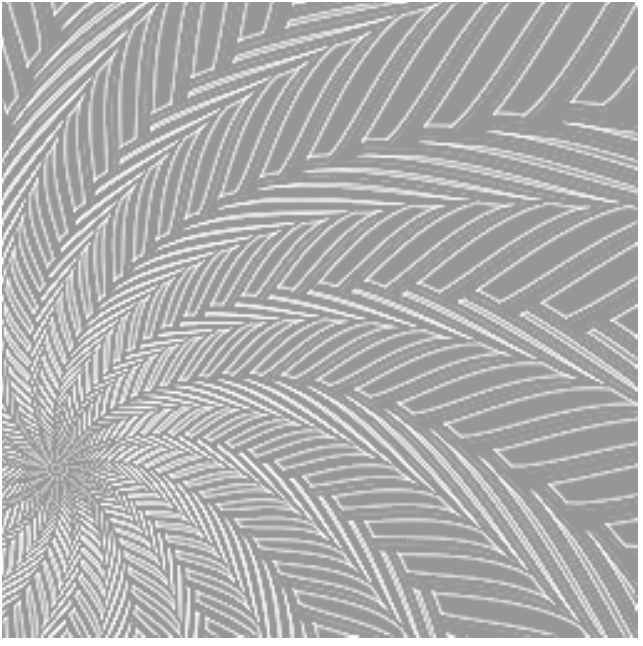} &
    \includegraphics[width=\sqimgwidth]{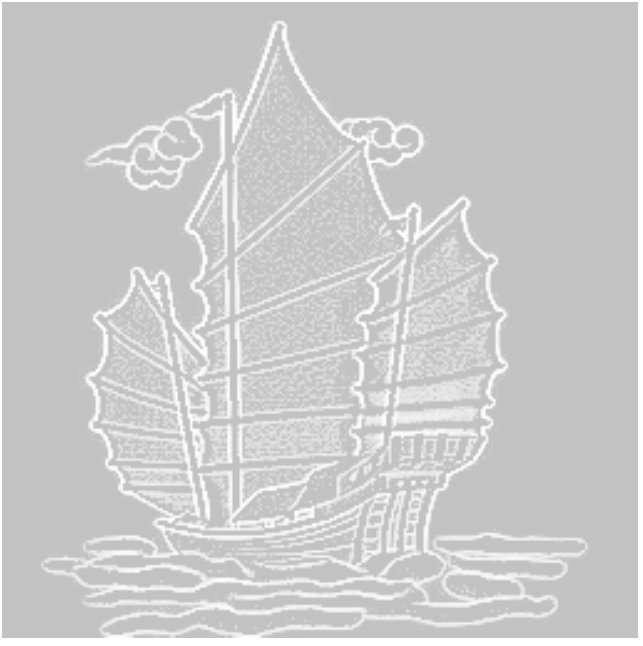} \\[-1.5mm]
    {\tiny TSVD 3826, {\bf SF 1569}} &
    {\tiny TSVD 2525, {\bf SF 1648}} &
    {\tiny TSVD 2907, {\bf SF 2032}} &
    {\tiny TSVD 7182, {\bf SF 5697}} &
    {\tiny TSVD 8629, {\bf SF 7058}} &
    {\tiny TSVD 2047, {\bf SF 1169}}
    \end{tabular}
    \caption{Example square matrices: (top) original images and (bottom) the gradient magnitude images (displayed after histogram equalization for better visibility). Image names are  approximation errors by using TSVD and SF with the same number of non-zeros are shown below the images. Boldface font indicates the winner for each case.}
    \label{fig:sqimgs}
\end{figure*}

There are many types of square matrices. TSVD is the best for approximating matrices close to low rank in terms of F-norm. For non-parametric SF, we performed an empirical study 1) to show that SF can supersede TSVD in F-norm by using the same number of non-zeros and 2) to identify the types of square matrices particularly suitable for SF.

Given a square matrix, we used the Matlab \texttt{fminunc} optimizer to solve the problem in Eq.~\ref{eq:np_sf}, where the non-zero entries in the factorizing matrices are initialized to random numbers between $[K^{-1}, K^{-1}+10^{-2}]$. The total number of non-zero entries for SF and TSVD are $N(\log_2 N)^2$ and $2Nr+r$. For a fair comparison, we set the $r=\lceil\log_2 N)^2/2\rceil$ in TSVD such that it has nearly the same number and no fewer non-zeros than SF.

We first used $256\times256$ grayscale images as approximated matrices because we can directly see them. Figure \ref{fig:sqimgs} (top) shows six typical square images, where we provide the resulting TSVD and SF approximation errors in F-norm below each image (more examples in Appendix \ref{appendix:non-param}).

The \texttt{chess} image (matrix) is close to low-rank because the black-and-white chessboard can is two-rank. Therefore TSVD performs better as expected. TSVD also works well for the \texttt{Lena} and \texttt{apple} images because TSVD tends to preserve low-frequency information in images. In contrast, TSVD is not as good as SF for the other three images that contain rich high-frequency details such as lines and corners, which indicates a matrix type where SF can defeat LRMF.

To further verify this, we computed the gradient magnitudes of the images (shown in Figure \ref{fig:sqimgs} bottom). In this way, the intensities in constant areas become zero, and the remaining non-zeros are mainly high-frequency details. We can see that SF gives a lower approximation error than TSVD for all such matrices, which confirms that SF is more advantageous for approximating matrices with rich high-frequency details.

In summary, the results show that the TSVD performance does not cap SF by using the same number of non-zeros for approximating square matrices. The winning cases indicate that SF is often better than LRMF when the approximated matrix is 1) sparse, 2) intrinsically high-rank, or 3) containing rich high-frequency details.

\setlength\tabcolsep{3pt}
\begin{table}[t]
\caption{Approximation errors in F-norm by using TSVD and SF for different types of square matrices.}
\label{tab:othermats}
\begin{center}
\begin{tabular}{llcc}
\hline\hline\\[-3.5mm]
Data type & Data name & TSVD & SF \\
\hline\\[-3mm]
 dense graph &        AuraSonar &   \bf{8.54e+00} &        8.68e+00 \\
 dense graph &          Protein &        1.17e+01 &   \bf{1.09e+01} \\
 dense graph &           Voting &   \bf{8.07e-04} &        1.71e+01 \\
 dense graph &            Yeast &        3.72e+01 &   \bf{3.61e+01} \\
     network &          Sawmill &        3.24e+00 &   \bf{1.03e+00} \\
     network &         Scotland &        5.90e+00 &   \bf{3.76e+00} \\
     network &             A99m &        1.47e+01 &   \bf{1.01e+01} \\
     network &    Mexican Power &        3.85e+00 &   \bf{1.71e+00} \\
     network &           Strike &        2.73e+00 &   \bf{1.04e+00} \\
     network &    Webkb Cornell &        6.98e+00 &   \bf{4.80e+00} \\
     network &       Worldtrade &        8.65e+04 &   \bf{4.47e+04} \\
surface mesh &          Mesh1e1 &        1.87e+01 &   \bf{9.82e+00} \\
surface mesh &          Mesh2e1 &   \bf{2.48e+02} &        3.47e+02 \\
surface mesh &     OrbitRaising &        9.37e+01 &   \bf{8.35e+01} \\
surface mesh &    Shuttle Entry &        2.73e+03 &   \bf{1.86e+03} \\
surface mesh & AntiAngiogenesis &        5.85e+01 &   \bf{3.29e+01} \\
  covariance &          Phoneme &   \bf{2.80e+01} &        5.27e+01 \\
  covariance &        MiniBooNE &   \bf{1.04e+00} &        6.36e+03 \\
  covariance &        Covertype &        8.22e-02 &   \bf{1.90e-02} \\
  covariance &            Mfeat &   \bf{1.11e+03} &        4.01e+05 \\
  covariance &        OptDigits &   \bf{3.28e+01} &        7.01e+01 \\
  covariance &        PenDigits &        4.00e+02 &   \bf{1.87e+02} \\
  covariance &         Acoustic &        1.36e-02 &   \bf{1.11e-02} \\
  covariance &            IJCNN &        5.24e-02 &   \bf{3.03e-02} \\
  covariance &         Spam Ham &        1.07e-01 &   \bf{4.97e-02} \\
  covariance &            TIMIT &   \bf{9.64e+01} &        1.56e+02 \\
  covariance &            Votes &        4.00e-01 &   \bf{1.70e-01} \\
\hline
\hline
\end{tabular}
\end{center}
\vspace{-3mm}
\end{table}

Besides square images, we have also compared TSVD and SF on several other types of square matrices. Table \ref{tab:othermats} shows the comparison results. The data types include affinity matrices of dense graphs (dense graph), affinity matrix of sparse networks (network), affinity matrix of surface mesh over 3D objects (surface mesh), and covariance matrix of vectorial data (covariance). We can see that for dense graph and covariance types, sometimes TSVD is better while sometimes SF can win. SF wins for most cases for surface mesh because the mesh networks probably do not have a low-rank structure. SF wins all data sets in the network type, which indicates that SF is more effective for approximating sparse matrices. We give the details of the data sets in Appendix \ref{appendix:details}.

\subsection{Approximation to Large Attention Matrices}
\label{sec:psfattn_synthetic}
Here we test whether PSF-Attn is scalable to approximate large attention matrices. We have used two synthetic data sets composed of long sequences for supervised learning tasks. A similar experimental setup appeared in \citep{hochreiter1997long} for scalability tests. The details of the data sets and tasks are given below:
\begin{itemize}
    \item \textit{Adding Problem}. This is a sequence regression task. Each element of an input sequence is a pair of numbers $(a_i, b_i)$, where $a_i\sim U(-1, 1)$, $b_i\in \{0, 1\}$, $i=1,\dots, N$. We generated signals at two randomly selected positions $t_1$ and $t_2$ such that $b_{t_1}=b_{t_2}=1$ and $b_i = 0$ elsewhere. The learning target is $y = 0.5+\dfrac{a_{t_1} + a_{t_2}}{4}$. For example, an input sequence $[(0.1, 0), (-0.4, 1), (0.3,0), (-0.2,0), (0.7,1)]$ will have the learning target $y=0.575$. Unlike \citep{hochreiter1997long}, we did not restrict the $t_1$ and $t_2$ choice to make the task more challenging. That is, the relevant signals can appear either locally or at a great distance from each other. In evaluation, a network prediction $\hat{y}$ is considered correct if $|y - \hat{y}| < 0.04$.
    \item \textit{Temporal Order}. This is a sequence classification task. A sequence consists of randomly chosen symbols from the alphabet $\{a, b, c, d, X, Y\}$, where the first four are noise symbols. Each sequence has two signal symbols, either $X$ or $Y$, which appear at two arbitrary positions. The four target classes correspond to the ordered combinations of the signal symbols $(X,X)$, $(X, Y)$, $(Y,X)$, and $(Y, Y)$. For example, an input sequence $[b,a,c,b,X,a,a,Y,b]$ should be classified as Class 2.
\end{itemize}
We prepared data of different sequence lengths for each problem. We started with $N=128$ and gradually increased the length by the factor of two, up to $N=2^{14}$. For every sequence length, we generated 200,000 training and 5,000 testing instances.

We compared PSF-Attn with several popular methods based on scaled dot-product attention (referred to as X-former architectures). The first two, Linformer \citep{linformer} and Performer \citep{performer} have also claimed to be scalable attention-based architectures. For completeness, we also included the original Transformer \citep{vaswani2017attention}. We have used their open-source PyTorch implementations\footnote{available at \url{https://github.com/lucidrains/linformer} and \url{https://github.com/lucidrains/performer-pytorch}}.

We followed standard cross-validation techniques to tune the main hyperparameters, such as the number of layers and heads, dimensionality of the token embedding, and query/key/value dimensions. For the Temporal Order problem, we directly fed the data instances to the embedding layers. For the Adding problem, the input data was only two-dimensional, and one of them was real-valued. Directly using such a low-dimensional embedding space would lead to poor attention. We augmented the dimensionality with a linear layer to assure sufficient freedom for the scaled dot-products in the X-former architectures. All the models were optimized using the Adam optimizer \citep{adam} with the learning rate of 0.001 using batch size of 40.

\newcommand{\synfigwidth}{7.5cm}
\begin{figure}[t]
	\begin{center}
		\includegraphics[width=\synfigwidth]{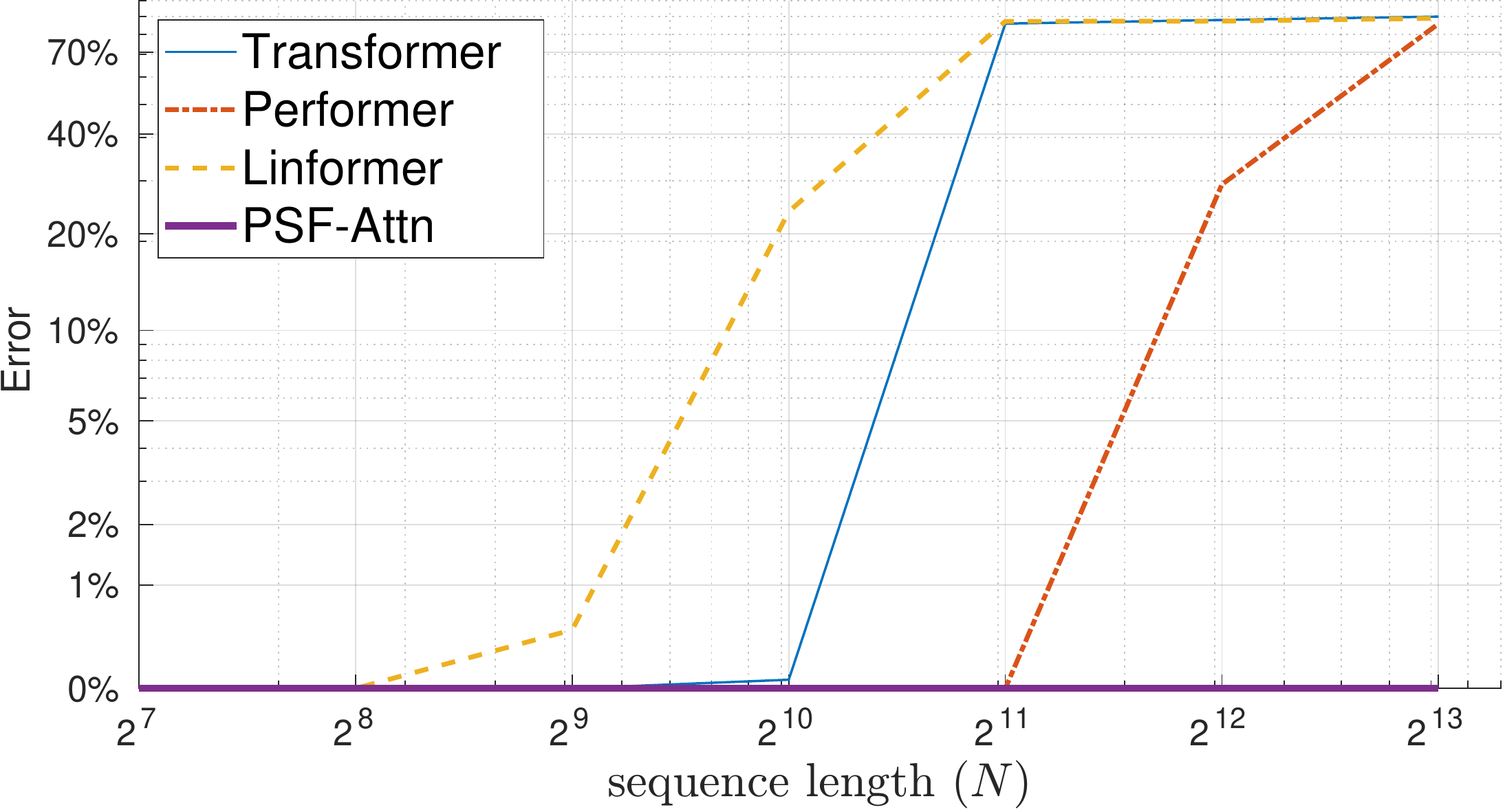}\quad
		\includegraphics[width=\synfigwidth]{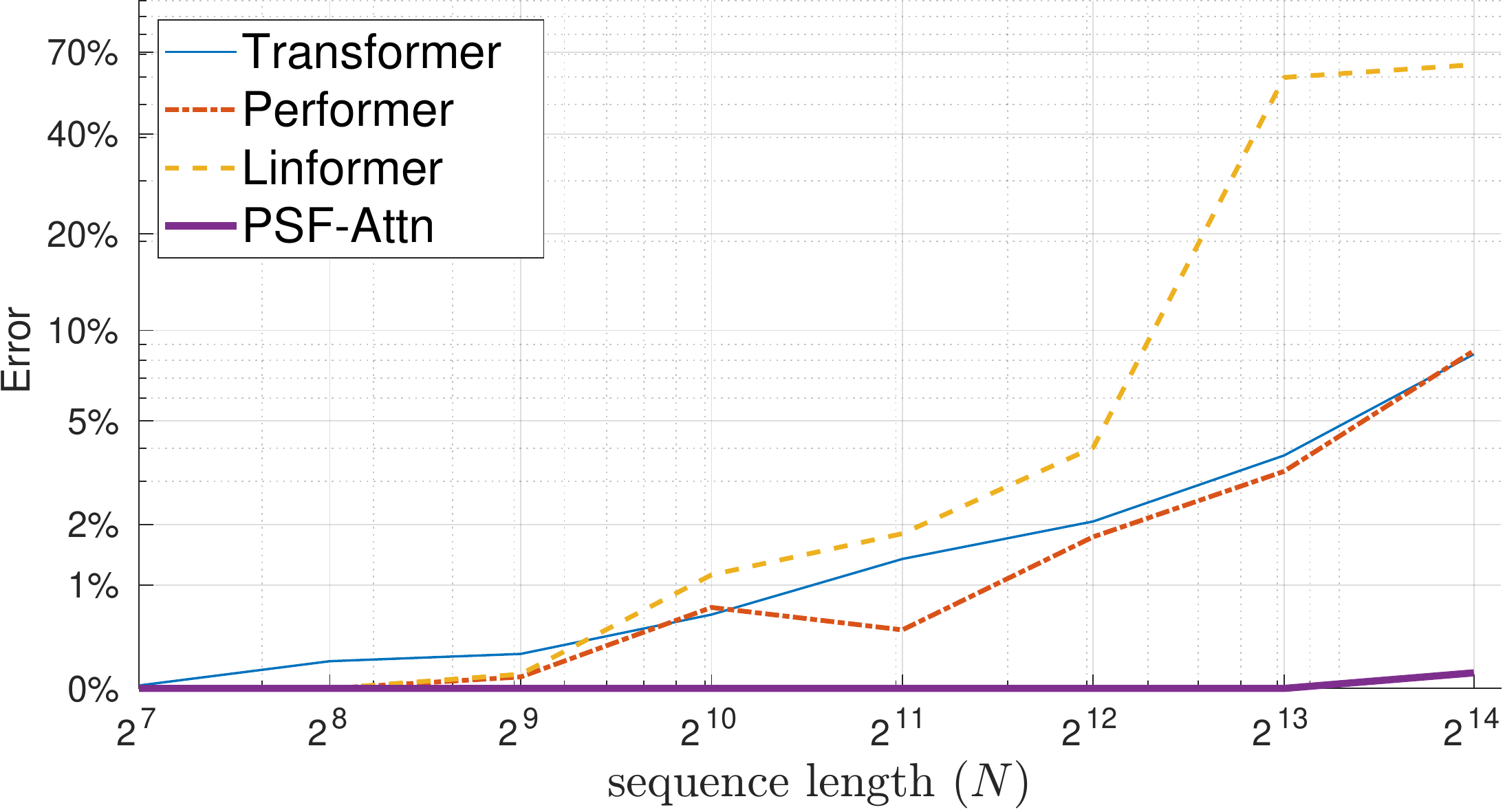}
	\end{center}
    \centering
    \caption{Error percentage of PSF-Attn and the X-formers for (left) the Adding problem and (right) the Temporal Order problem with increasing sequence lengths.}
    \label{fig:synthetic}
    \vspace{-3mm}
\end{figure}

The results are shown in Figure \ref{fig:synthetic}. For the Adding problem, we see that all models work fine for short sequences ($100\%$ for $N\leq 256$). The X-former methods, however, turn worse or even useless when the sequences are longer. Linformer has an error rate of $0.48\%$ for $N=512$ and $86.18\%$ for $N=1024$, which is nearly as bad as random guessing ($92\%$). Transformer becomes problematic ($84.48\%$ error) when $N\geq2048$. Performer starts to get wrong when $N=4096$, giving only $71.76\%$ accuracy, and when $N=8192$, its prediction becomes almost random guessing. In contrast, PSF-Attn achieves $100\%$ accuracy for all the tested lengths.

A similar pattern holds for the Temporal Order problem. When $N\leq 512$, all compared models give perfect or nearly perfect predictions. With longer sequences, for example when $N=4096$, the error rates of Transformer, Performer, and Linformer become $2.06\%$, $1.76\%$, and $4.02\%$, respectively. When $N=16324$, their error rates become $8.38\%$, $8.60\%$, and $64.22\%$, respectively. In contrast, PSF-Attn achieves $100\%$ accuracy for $N\leq 8192$, and $99.89\%$ accuracy for $N=16324$.

In summary, our method has better scalability than the three attention models based on scaled dot-products in terms of lower learning errors. PSF-Attn can still achieve nearly $100\%$ prediction accuracy for an attention matrix size up to tens of thousands.

\subsection{Long Range Arena Public Benchmark}
\label{sec:psfattn_realworld}

\setlength\tabcolsep{1pt}
\newcommand{\lraegfigwidth}{2.6cm}
\begin{figure}[t]
	\begin{center}
	\begin{tabular}{cccc}
		\includegraphics[width=2.55cm]{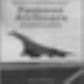} &
		\includegraphics[width=2.55cm]{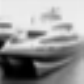} &
		\includegraphics[width=2.6cm]{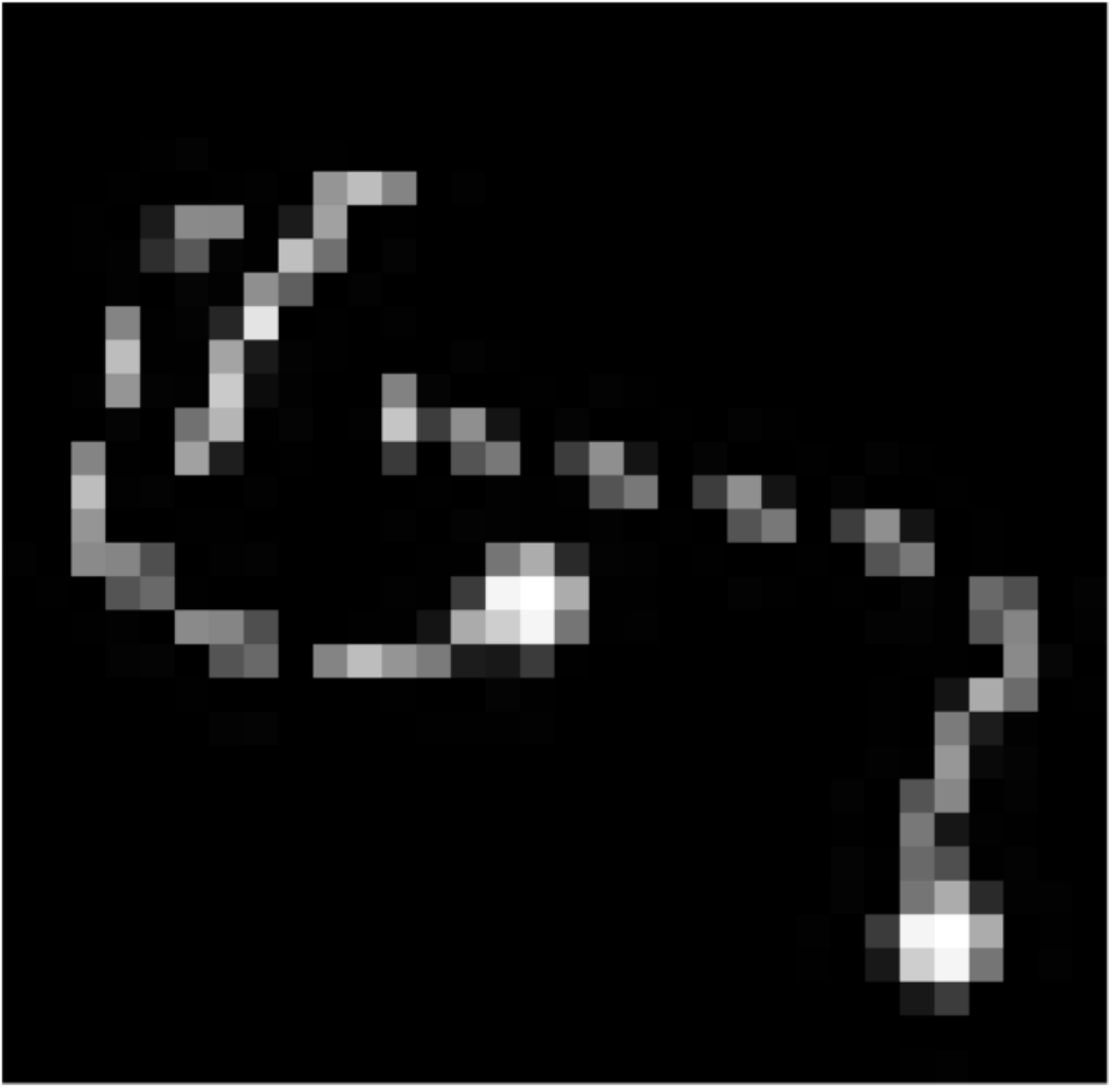} &
		\includegraphics[width=2.6cm]{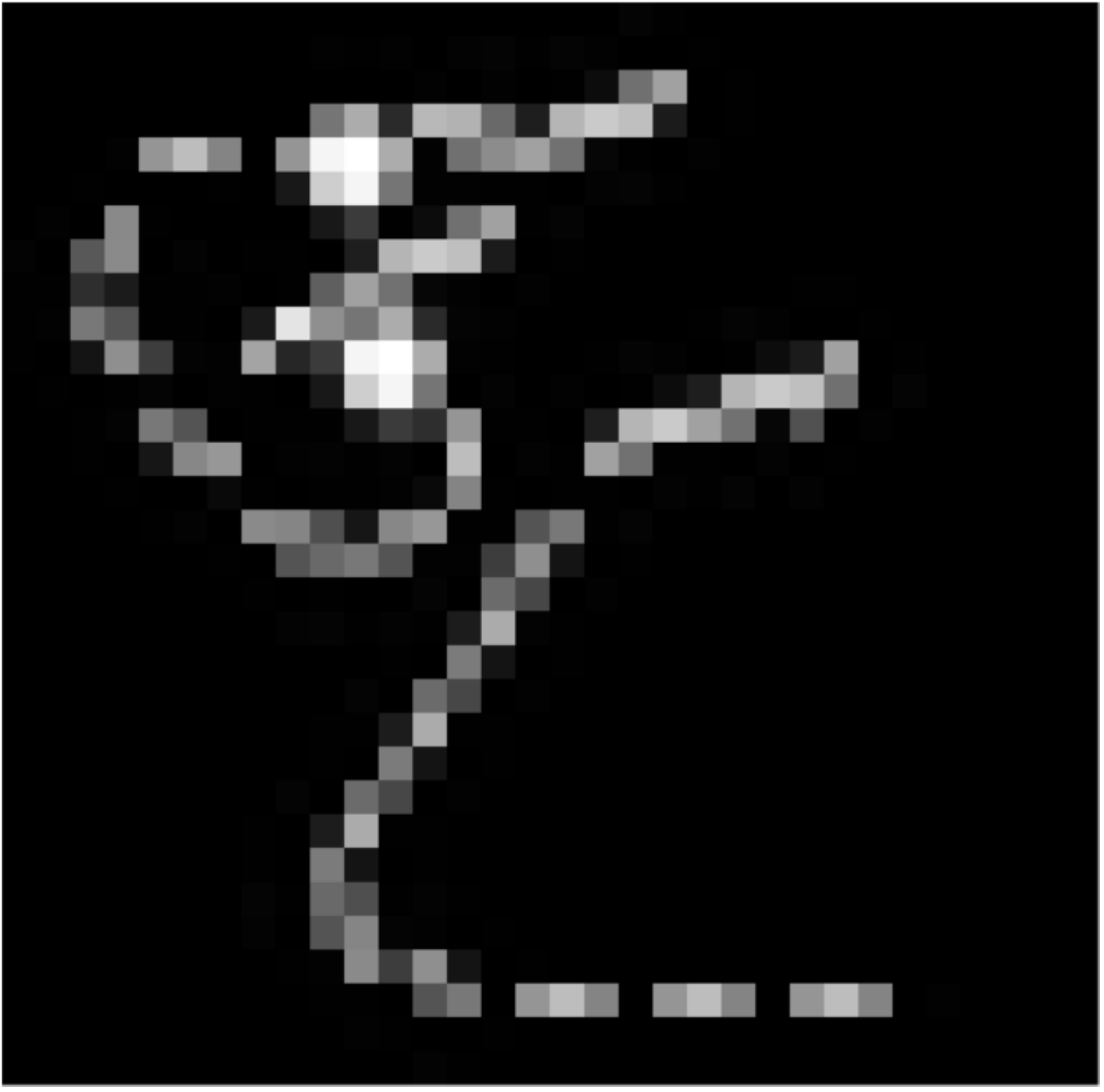}\\
		{\tiny {\bf Image, airplane}} &
		{\tiny {\bf Image, ship}}&
		{\tiny {\bf Pathfinder, Negative}} & 
		{\tiny {\bf Pathfinder, Positive}}\\
 	\end{tabular}
	\end{center}
	\caption{Example matrices: (left two) from the Image Classification and (right two) from Pathfinder tasks.}
	\label{fig:lra}
\end{figure}

Next we provide the experimental results on Long Range Arena (LRA), a publicly available benchmark for modeling long sequential data \citep{tay2020long}. We select four tasks from LRA that cover various data types and demand flexible reasoning abilities of tested models. 

\setlength\tabcolsep{5pt}
\begin{table*}[t]
\centering
\caption{Classification accuracies by the compared methods for the four LRA tasks. A dash (``-'') means the result is absent in the corresponding paper. For PSF-Attn, we present the mean ($\mu$) and standard deviation ($\sigma$) across multiple runs in the $\mu\pm\sigma$ format.}
\label{tab:LRAacc}
\begin{tabular}{lcccc}
\hline\hline
Model              & ListOps & Text & Image & Pathfinder\\
                   & $N=2000$ & $N=4000$ & $N=1024$ & $N=1024$\\
\hline
Transformer \citep{tay2020long}    & 36.37 & 64.27 & 42.44 & 71.40\\
Transformer  \citep{zhu2021long}   & 37.13 & 65.35 & -     & -    \\
Transformer \citep{xiong2021nystr}    & 37.10 & 65.02 & 38.20 & 74.16\\
\hline
Sparse Transformer \citep{tay2020long}   & 17.07 & 63.58 & 44.24 & 71.71\\
\hline
Longformer \citep{tay2020long}     & 35.63 & 62.58 & 42.22 & 69.71\\
\hline
Linformer \citep{tay2020long}      & 37.70 & 53.94 & 38.56 & 76.34\\
Linformer \citep{zhu2021long}      & 37.38 & 56.12 & -     & -    \\
Linformer \citep{xiong2021nystr}      & 37.25 & 55.91 & 37.84 & 67.60\\
\hline
Reformer \citep{tay2020long}       & 37.27 & 56.10 & 38.07 & 68.50\\
Reformer \citep{zhu2021long}       & 36.44 & 64.88 & -     & -    \\
Reformer \citep{xiong2021nystr}       & 19.05 & 64.88 & 43.29 & 69.36\\
\hline
Performer \citep{tay2020long}      & 18.01 & 65.40 & 42.77 & 77.05\\
Performer \citep{zhu2021long}      & 32.78 & 65.21 & -     & -    \\
Performer \citep{xiong2021nystr}      & 18.80 & 63.81 & 37.07 & 69.87\\
\hline
BigBird \citep{tay2020long}        & 36.06 & 64.02 & 40.83 & 74.87\\
\hline
Linear Transformer \citep{tay2020long}  & 16.13 & 65.90 & 42.34 & 75.30\\
\hline
Transformer-LS \citep{zhu2021long} & 38.36 & 68.40 & -     & -    \\
\hline
RFA-Gaussian \citep{peng2021random}   & 36.80 & 66.00 & -     & -    \\
\hline
Nystr\"omformer \citep{zhu2021long}  & 37.34 & 65.75 & -     & -    \\
Nystr\"omformer \citep{xiong2021nystr}  & 37.15 & 65.52 & 41.58 & 70.94\\
\hline
PSF-Attn & \textbf{38.85}$\pm$0.06 & \textbf{77.32}$\pm$0.25 &  \textbf{45.01}$\pm$0.21 & \textbf{80.49}$\pm$0.13\\
\hline
\hline
\end{tabular}
\vspace{-3mm}
\end{table*}

\begin{itemize}
    \item \emph{ListOps}. This is a sequence classification task for measuring the ability of models to identify and parse hierarchically constructed data \citep{nangia2018listops}. We used an enlarged version of the original ListOps, with a max sequence length up to 2000 and tree depth up to 10 \citep{tay2020long}. Each element in a sequence can be an operator, a digit, and a left or right bracket. The brackets define lists of items. Each operator, MAX, MIN, MED, and SUM\_MOD, takes the items in a list as input and returns a digit, where MED means median and SUM\_MOD means summation followed by modulo 10. An example sequence [MAX 6 [MED 3 2 2] 8 5 [MIN 8 6 2]] has ground truth answer 8. A prediction is correct if an output value of a neural network matches the ground truth label. For good predictions, a model should access all sequence elements and identify the proper parsing structure.
    \item \emph{Text Classification} This is a binary sentiment classification task constructed from the IMDb reviews \citep{maas2011learning}. Given a review as a sequence of characters, the goal is to classify it as positive or negative. Due to the character-level representation, the sequences are much longer than the word-level version used in conventional language modeling. We truncated or padded every sequence to a fixed length ($N=4000$).
    \item \emph{Image Classification}. This task is to classify images into one of ten classes. The images and class labels come from the grayscale version of CIFAR10. Each image is flattened to form a sequence of length 1024. Unlike conventional computer vision, the task requires the predictors to treat the grayscale levels (0-255) as categorical values. That is, each image becomes a sequence of symbols with an alphabet size of 256. Two example images and their class labels are shown in Figure \ref{fig:lra}.
    \item \emph{Pathfinder}. This is a binary classification task on synthetic images, which is motivated by cognitive psychology \citep{linsley2018learning}. Each image (size $32\times 32$) contains two highlighted endpoints and some path-like patterns. Similar to the Image Classification task, the predictors must treat the pixels as categorical values and flatten the image to a sequence of length 1024. The task is to classify whether there is a path consisting of dashes between two highlighted points. Two example images and their classes are shown in Figure \ref{fig:lra}.
\end{itemize}

\begin{figure*}[t]
	\begin{center}
		\includegraphics[width=0.7\textwidth]{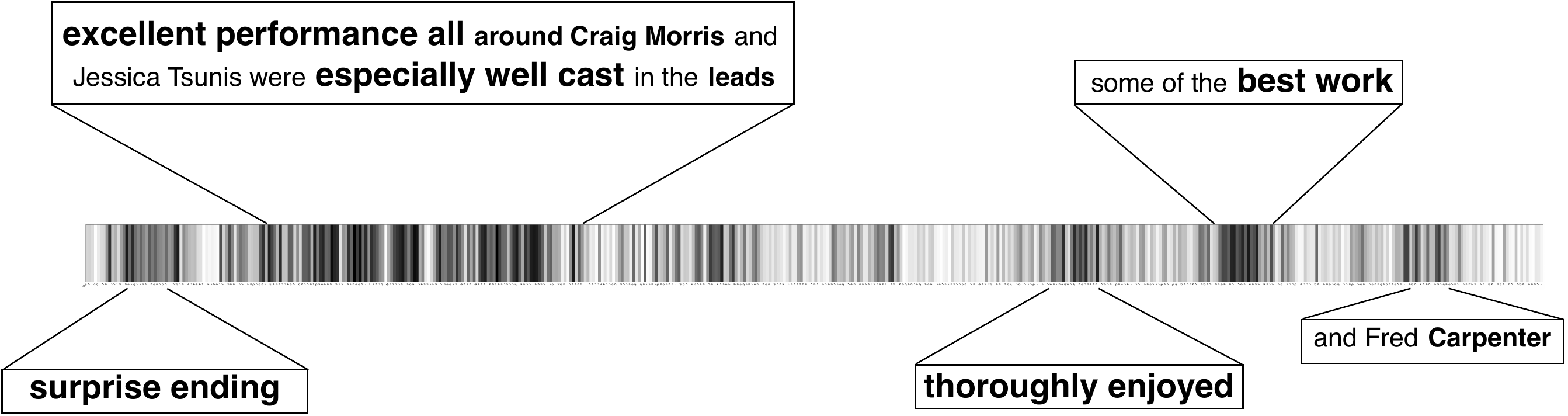}
	\end{center}
	\caption{Attention vector visualization of a positive review in Text Classification. Darker cells have larger absolute values.}
	\label{fig:attn_IMDb}
	\vspace{-3mm}
\end{figure*}

\newcommand{\attmapfigwidth}{3cm}
\begin{figure}[t]
	\begin{center}
		\includegraphics[width=\attmapfigwidth]{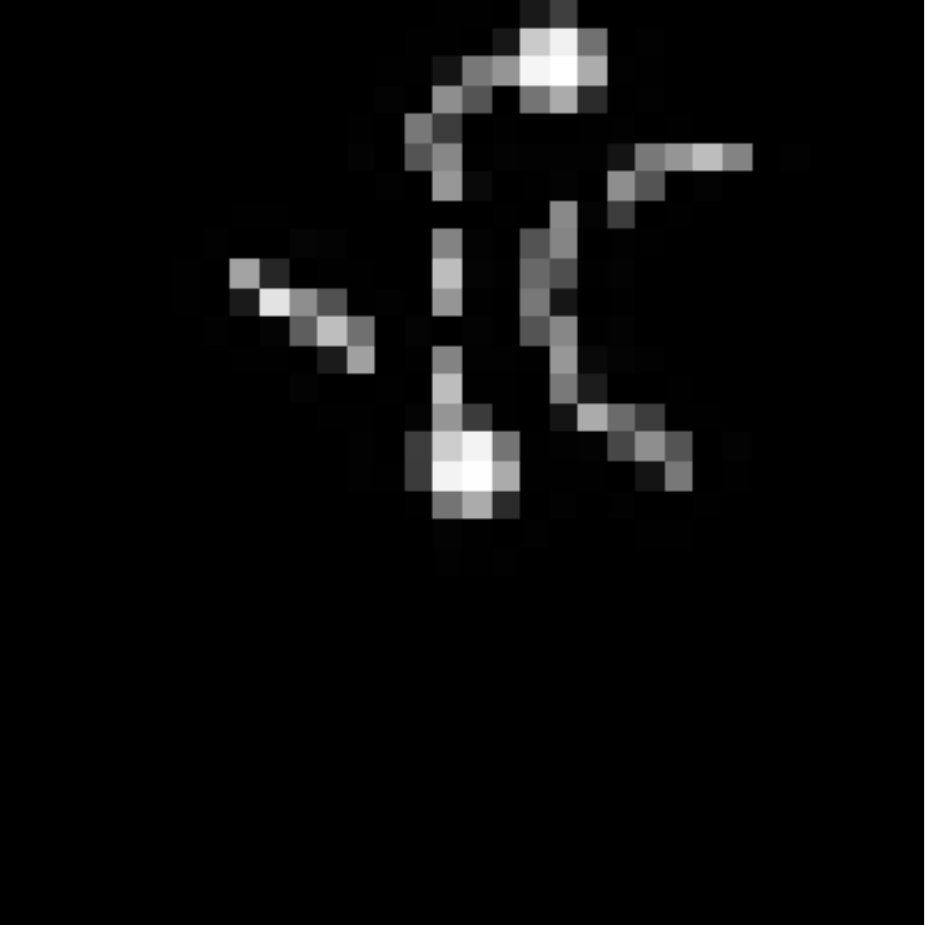}\quad
		\includegraphics[width=3.666cm]{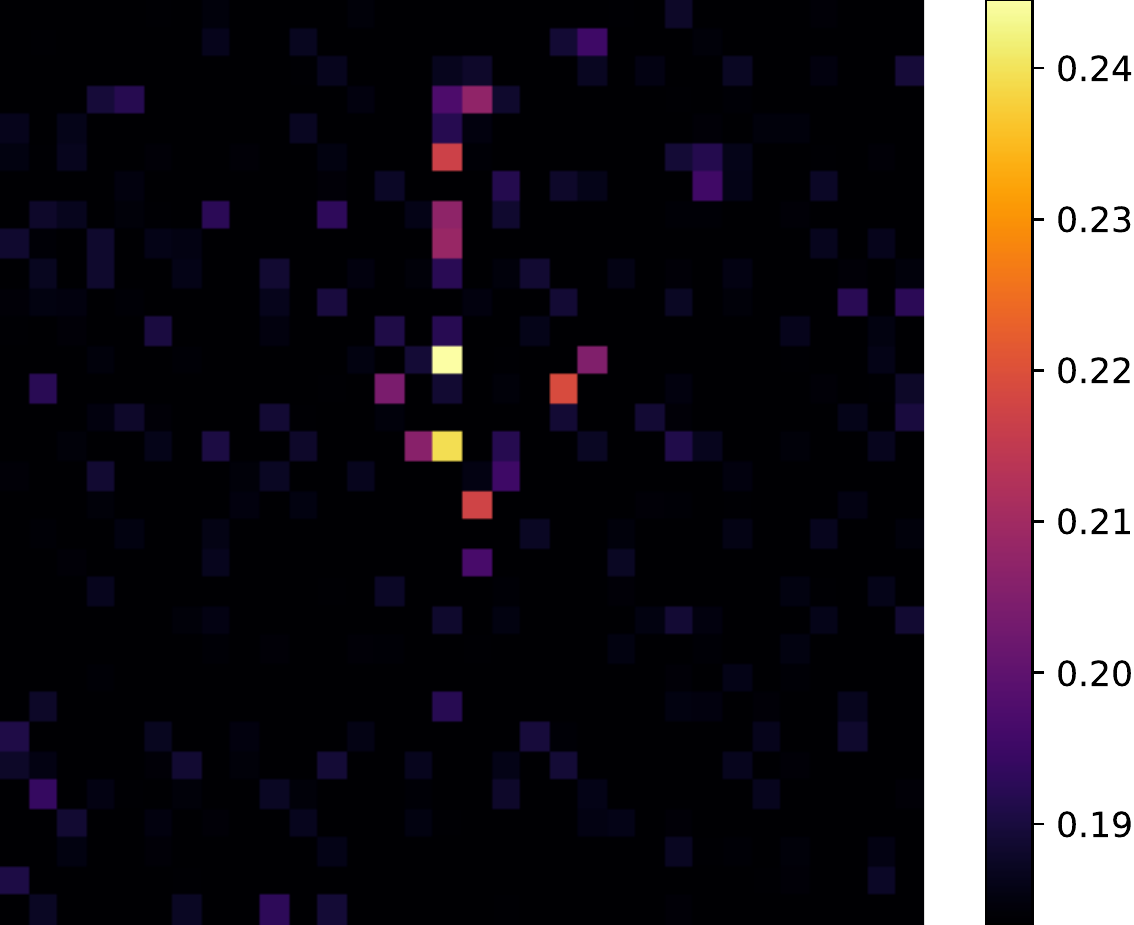}
	\end{center}
	\vspace{-2mm}
	\caption{Visualization of (left) a positive Pathfinder instance and (right) its attention values.}
	\label{fig:attn_pathfinder}
	\vspace{-3mm}
\end{figure}

We have tested PSF-Attn in the above LRA learning tasks, where the hyperparameters in PSF-Attn were again tuned by cross-validation. No external data was used for pre-training. We ran PSF-Attn four times with a different random seed for each task. The mean and standard deviation across the multiple runs are reported in Table \ref{tab:LRAacc}.

For comparison, we quote the prediction accuracies reported for many X-former methods in the literature, including Transformer \citep{transformer}, Sparse Transformer \citep{child2019generating}, Longformer \citep{beltagy2020longformer}, Linformer \citep{linformer}, Reformer \citep{kitaev2020reformer}, Performer \citep{performer}, BigBird \citep{zaheer2020big}, Linear transformer \citep{katharopoulos2020transformers}, Transformer-LS \citep{zhu2021long}, RFA-Gaussian \citep{peng2021random}, and Nystr\"omformer \citep{xiong2021nystr}. If a method has different implementations, we quote all variants and their results. We exclude results that rely on external data for pre-training.

We see that PSF-Attn wins all tasks by giving the best classification accuracy among all compared methods. Such strong cross-task wins suggest that our method usually provides better attention approximation than those based on scaled dot-products. 

Remarkably, our method has substantially improved the state-of-the-art in the Text and Pathfinder tasks. For Text, PSF-Attn achieves $77.32\%$ accuracy, compared to the runner-up $68.4\%$ by Transformer-LS. Our method wins by $80.49\%$ accuracy for Pathfinder, which gains about $5\%$ higher than the best X-former (Linear Transformer $75.3\%$). The significant improvement brought by PSF-Attn is probably because the two tasks involve sparse attention matrices.

We also investigated whether the approximating attention $\Xh=\prod_mW^{(m)}$ is meaningful by visualization.
Figure \ref{fig:attn_IMDb} shows an attention vector (absolute values) of token [``CLS'']. Tokens in the review having more weight are highlighted. We see that PSF-Attn has a good performance in capturing sparse attention and identify the relevant words.

Figure \ref{fig:attn_pathfinder} illustrates the attention values for a positive instance in Pathfinder. We calculated the attention vector of the $i$-th point as the $i$-th row in $\Xh$. We then summed the attention vectors of the endpoints and their direct neighbors and reshaped the sum to $32\times 32$ for visualization, where we see that the positions with high absolute visualized values distribute around the connecting path between the endpoints.

\section{Conclusion}
\label{sec:conclusion}
We have proposed a new approximation method called Sparse Factorization for square matrices by using a product of sparse matrices. Given the same budget of non-zero numbers, our method has shown superior performance over conventional low-rank matrix factorization approaches, especially in cases where the approximated matrix is sparse, high-rank, or contains high-frequency details. We have also given parametric design and performed an empirical study on the classification of long sequential data. As the critical attention component, our method has demonstrated clear wins over the conventional scaled dot-product transformer and its several variants in terms of scalability and accuracy.

We have employed the Chord protocol to fix the non-zero positions in the sparse factorizing matrices. Later we could consider the other predefined protocols or even adaptively learned protocols for the sparse structure. In this work, we have considered approximating unconstrained square matrices. In the future, we could investigate the approximation to more specific matrices such as non-negative, stochastic, symmetric, or semi-definite matrices. As a result, we could extend the method for further applications such as large-scale graph matching, Gaussian Process, or network structure identification.


\bibliographystyle{plainnat}

\clearpage
\begin{center}
\textbf{\large Appendix. Sparse Factorization of Large Square Matrices}
\end{center}
\setcounter{equation}{0}
\setcounter{figure}{0}
\setcounter{table}{0}
\setcounter{page}{1}
\setcounter{section}{0}
\makeatletter
\renewcommand{\theequation}{A\arabic{equation}}
\renewcommand{\thefigure}{A\arabic{figure}}
\renewcommand{\thetable}{A\arabic{table}}

\section{Non-parametric Experiments}
\begin{figure*}[t]
	\centering
	\begin{tabular}{cccccc}
		abstract & bridge & upper & middle & bottom & moon \\
		\includegraphics[width=\sqimgwidth]{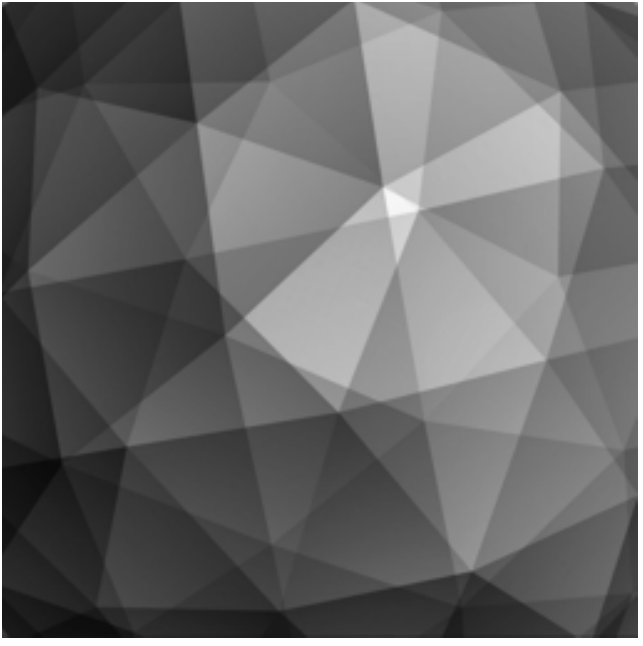} &
		\includegraphics[width=\sqimgwidth]{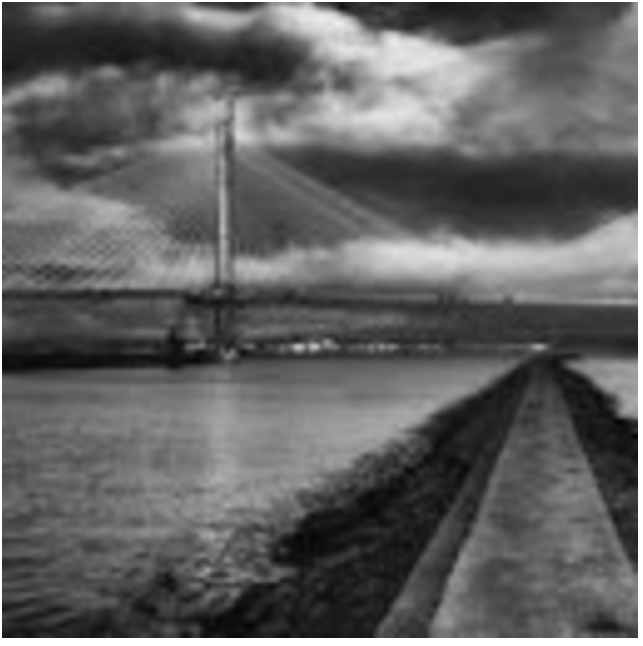} &
		\includegraphics[width=\sqimgwidth]{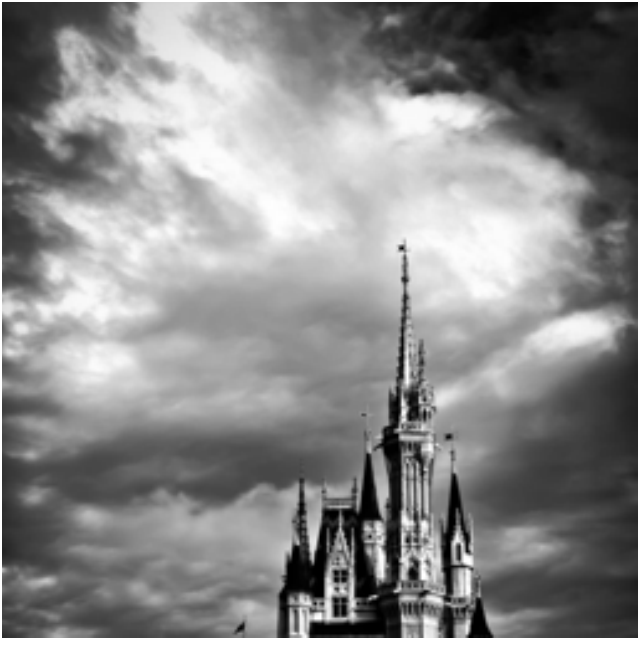} &
		\includegraphics[width=\sqimgwidth]{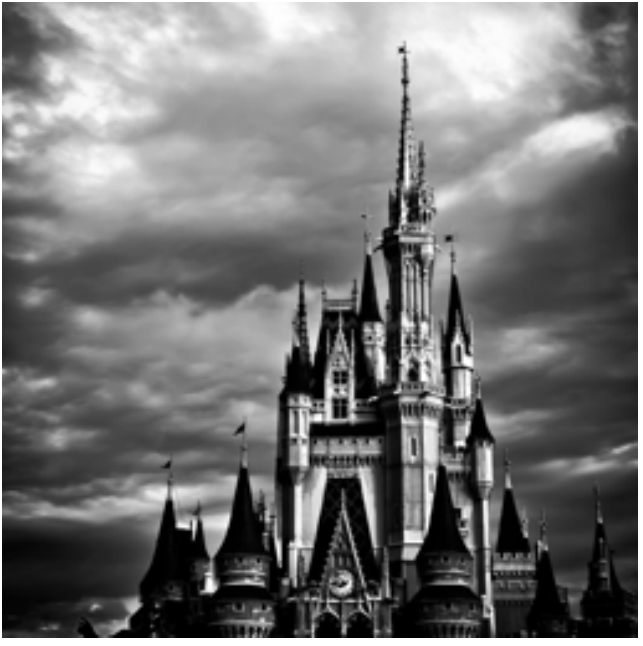} &
		\includegraphics[width=\sqimgwidth]{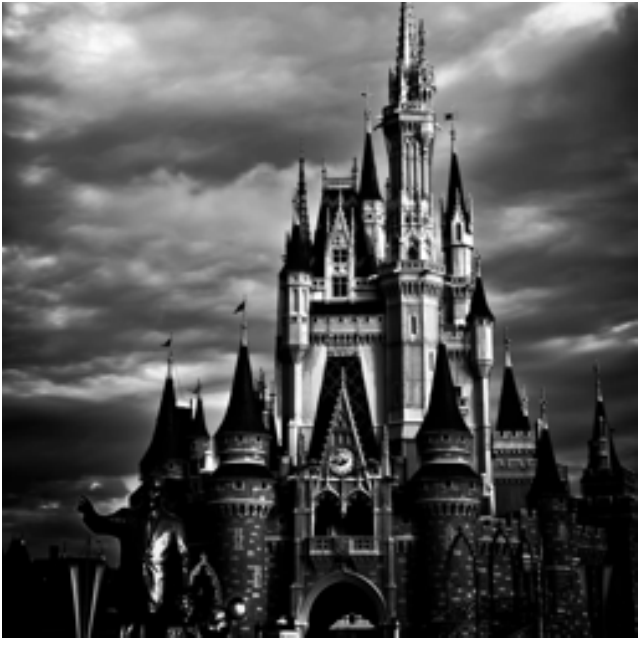} &
		\includegraphics[width=\sqimgwidth]{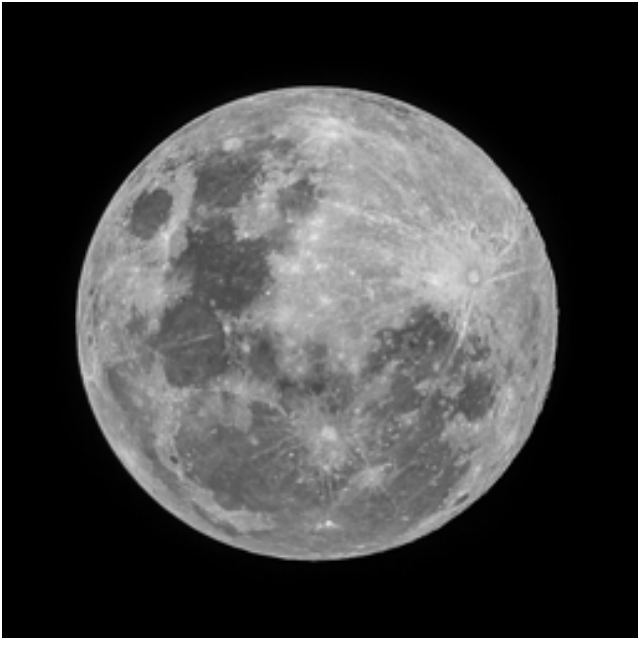} \\[-1.5mm]
		{\tiny {\bf TSVD 820}, SF 1312} &
		{\tiny {\bf TSVD 1008}, SF 1308} &
		{\tiny {\bf TSVD 1671}, SF 2378} &
		{\tiny {\bf TSVD 2623},  SF 2879} &
		{\tiny TSVD 3121, {\bf SF 2863}} &
		{\tiny TSVD 1655, {\bf SF 1462}}\\[2mm]
		\includegraphics[width=\sqimgwidth]{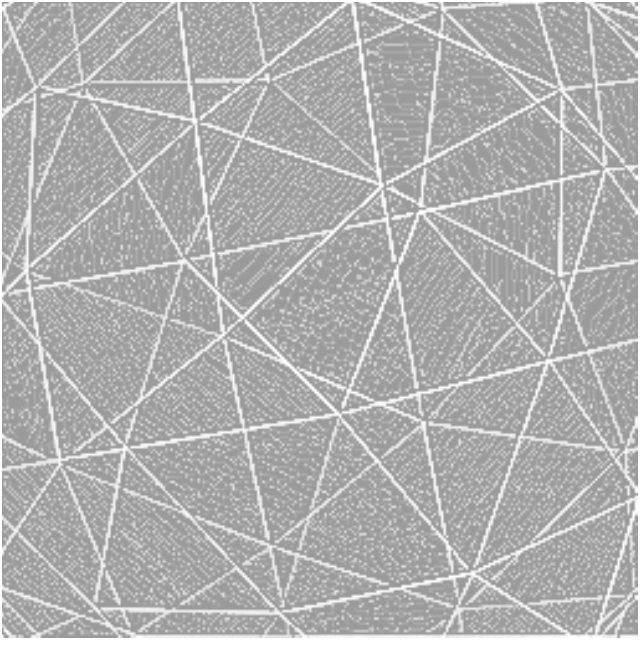} &
		\includegraphics[width=\sqimgwidth]{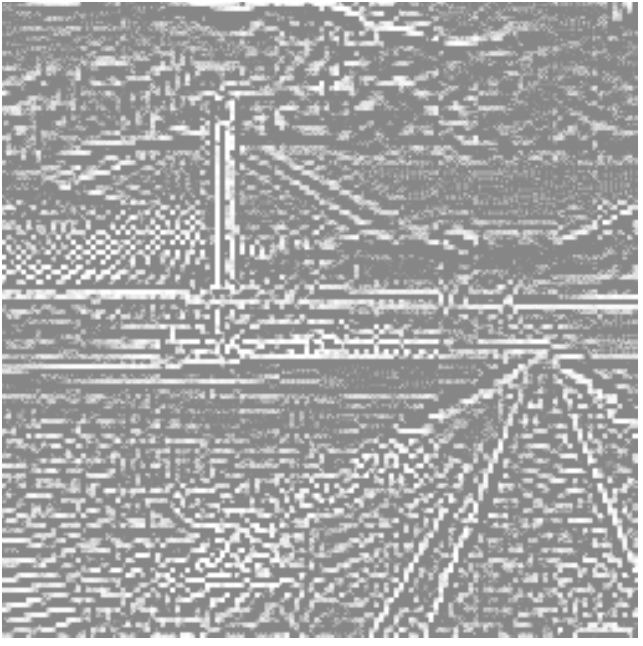} &
		\includegraphics[width=\sqimgwidth]{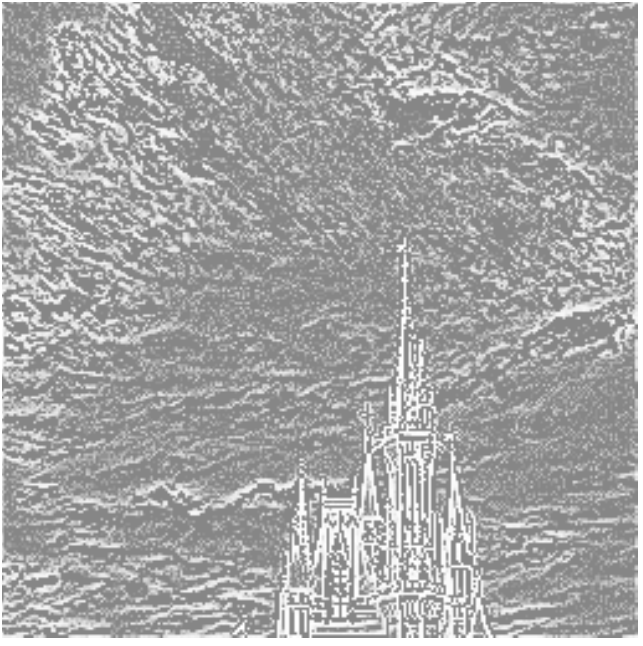} &
		\includegraphics[width=\sqimgwidth]{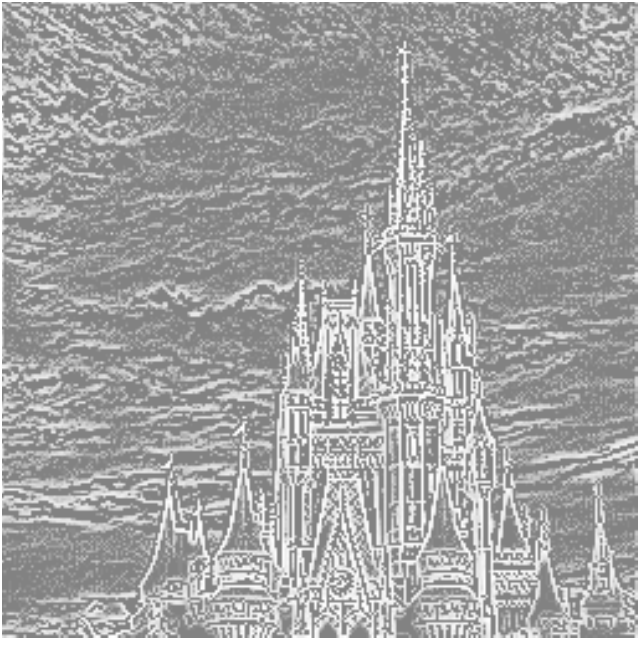} &
		\includegraphics[width=\sqimgwidth]{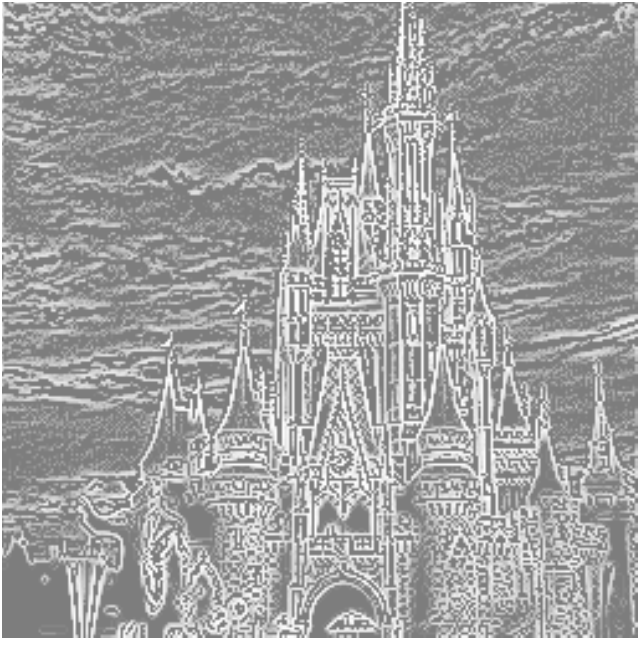} &
		\includegraphics[width=\sqimgwidth]{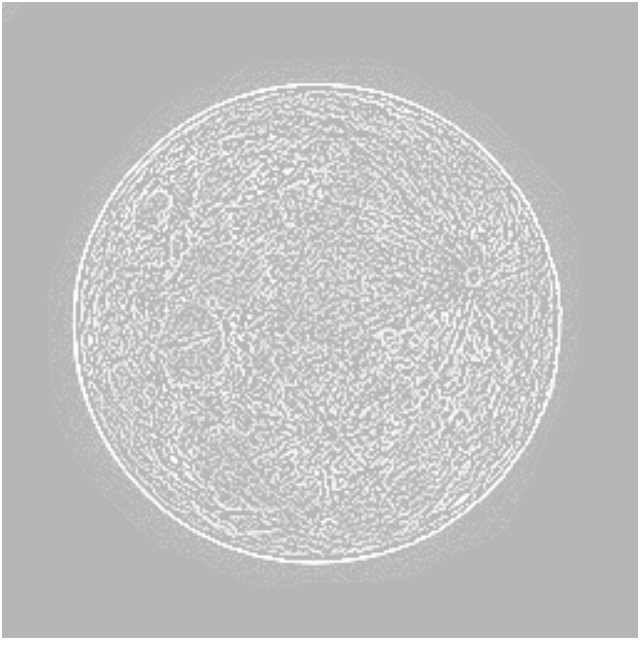} \\[-1.5mm]
		{\tiny TSVD 1079, {\bf SF 839}} &
		{\tiny TSVD 679, {\bf SF 401}} &
		{\tiny TSVD 1824, {\bf SF 865}} &
		{\tiny TSVD 3612, {\bf SF 1673}} &
		{\tiny TSVD 4344, {\bf SF 2656}} &
		{\tiny TSVD 2743, {\bf SF 1666}}
	\end{tabular}
\caption{Other example square matrices: (top) original images and (bottom) the gradient magnitude images (displayed after histogram equalization for better visibility). Image names are  approximation errors by using TSVD and SF with the same number of non-zeros are shown below the images. Boldface font indicates the winner for each case.}
\label{fig:othersqimgs}
\end{figure*}

In addition to six square images from the main paper, we have studied another six square images using the previously described approach. The results are shown in Figure \ref{fig:othersqimgs}. The conclusions are aligned with those in the main paper
\begin{itemize}
    \item There exist cases where SF performs better than TSVD, which means the SF approximation quality is not capped by TSVD by using the same number of non-zeros.
    \item SF is more likely to win for images where there are more high-frequency details, especially for the gradient-magnitude images.
\end{itemize}
\label{appendix:non-param}

\subsection{Details of Other Square Matrices}
\label{appendix:details}
In the main paper Table 1, we have present comparison between TSVD and SF for approximating different types of square matrices using the same number of non-zeros. Here we provide the sources and statistics of the square matrices.
\begin{itemize}
    \item \texttt{AuralSonar} ($N=100$). It is the \emph{Aural Sonar} data set from \citep{chen2009similarity}. The original research investigated the human ability to distinguish different types of sonar signals by ear (Philips et al., 2006).
    \item \texttt{Protein} ($N=213$). It is the \emph{Protein} data set from \citep{chen2009similarity}, which contains the radial basis function (RBF) kernel between 213 proteins.
    \item \texttt{Voting} ($N=435$). It is the \emph{Voting} data set from \citep{chen2009similarity}, which contains dissimilarities between 435 voting records with 16 scaled voting attributes.
    \item \texttt{Yeast} ($N=200$). It is the \emph{Yeast-SW-7-12} data set from the same repository in \citep{chen2009similarity}. The data set converts the pairwise Smith-Waterman similarities $s_{ij} $\citep{lanckriet2004kernel,xu2014ricci} to dissimilarities by $d_{ij}=\sqrt{s_{ii}+s_{jj}-s_{ij}-s_{ji}}$.
    \item \texttt{Sawmill} ($N=36$). It is the \emph{Sawmill} communication network from the Pajek data sets\footnote{available at \url{http://vlado.fmf.uni-lj.si/pub/networks/data/}}. This is a sparse matrix with 124 non-zero entries.
    \item \texttt{Scotland} ($N=108$). It is the \emph{Scotland} network from the Pajek data sets, which is about Corporate interlocks in Scotland (1904-5). The matrix is sparse with 644 non-zero entries.
    \item \texttt{A99m} ($N=234$). It is the \emph{GD'99 - Linden strasse} network from the Pajek data sets. The network is about the characters and their relations in the long-running German soap opera called `Lindenstrasse'. The matrix is sparse with 510 non-zero entries.
    \item \texttt{Mexican power} ($N=35$). It is the \emph{Mexican} network from the Pajek data sets. The network contains the core of this political elite: the presidents and their closest collaborators. The matrix is sparse with 117 non-zero entries.
    \item \texttt{Strike} ($N=24$). It is the \emph{Strike} network from the Pajek data sets. This is a social network about informal communication within a sawmill on strike. The matrix is sparse with 38 non-zero entries.
    \item \texttt{Webkb Cornell} ($N=195$). It is the Cornell subset in the LINQS WebKB data set\footnote{available at \url{https://linqs.soe.ucsc.edu/data}}. The network is about citations among the 195 publication from Cornell. The matrix is sparse with 304 non-zero entries.
    \item \texttt{WorldTrade} ($N=80$). It is the \emph{World\_trade} network in the Pajek data sets. The network is about world trade in miscellaneous manufactures of metal, 1994. The matrix is sparse with 998 non-zero entries.
    \item \texttt{Mesh1e1} ($N=48$). It is the \emph{mesh1e1} data set in SuiteSparse Matrix Collection\footnote{available at \url{https://sparse.tamu.edu/}}. The matrix was originally from NASA, collected by Alex Pothen. It is a sparse matrix with 306 non-zero entries.
    \item \texttt{Mesh2e1} ($N=306$). It is the \emph{mesh2e1} data set in SuiteSparse Matrix Collection. The matrix was originally from NASA, collected by Alex Pothen. It is a sparse matrix with 2018 non-zero entries.
    \item \texttt{OrbitRaising} ($N=442$). It is the \emph{orbitRaising\_1} data set in SuiteSparse Matrix Collection. The matrix was from an optimal control problem. It is a sparse matrix with 2906 non-zero entries.
    \item \texttt{Shuttle Entry} ($N=560$). It is the \emph{spaceShuttleEntry\_1} data set in SuiteSparse Matrix Collection. The matrix was from an optimal control problem. It is a sparse matrix with 6891 non-zero entries.
    \item \texttt{AntiAngiogenesis} ($N=205$). It is the \emph{tumorAntiAngiogenesi\_1} data set in SuiteSparse Matrix Collection. The matrix was from an optimal control problem. It is a sparse matrix with 1783 non-zero entries.
    \item \texttt{Phoneme} ($N=256$). It is the covariance matrix of the \emph{Phoneme} data set accompanied with the Elements of Machine Learning book \citep{hastie01statisticallearning}. The original data\footnote{available at \url{https://web.stanford.edu/~hastie/ElemStatLearn/data.html}} has 4508 instances of 256 dimensions.
    \item \texttt{MiniBooNE} ($N=50$). It is the covariance matrix of the \emph{MiniBooNE particle identification} data set in the UCI Repository\footnote{available at \url{https://archive.ics.uci.edu/ml/}}. The original data has 130064 instances of 50 dimensions.
    \item \texttt{Covertype} ($N=54$). It is the covariance matrix of the \emph{Covertype} data set in the UCI Repository. The original data has 581012 instances of 54 dimensions.
    \item \texttt{Mfeat} ($N=649$). It is the covariance matrix of the \emph{Multiple Features} data set in the UCI Repository. The original data has 2000 instances of 649 dimensions.
    \item \texttt{OptDigits} ($N=64$). It is the covariance matrix of the \emph{Optical Recognition of Handwritten Digits} data set in the UCI Repository. The original data has 5620 instances of 64 dimensions.
    \item \texttt{PenDigits} ($N=16$). It is the covariance matrix of the \emph{Pen-Based Recognition of Handwritten Digits} data set in the UCI Repository. The original data has 10992 instances of 16 dimensions.
    \item \texttt{Acoustic} ($N=50$). It is the covariance matrix of the \emph{SensIT Vehicle (acoustic)} data set in the LIBSVM Classification (Multi-class) data collection\footnote{available at \url{https://www.csie.ntu.edu.tw/~cjlin/libsvmtools/datasets/multiclass.html}}. The original data has 98528 instances of 50 dimensions.
    \item \texttt{IJCNN} ($N=22$). It is the covariance matrix of the \emph{ijcnn1} data set in the LIBSVM Classification (Binary class) data collection\footnote{available at \url{https://www.csie.ntu.edu.tw/~cjlin/libsvmtools/datasets/binary.html}}. The original data has 126701 instances of 22 dimensions.
    \item \texttt{Spam Ham} ($N=448$). It is the covariance matrix of a data set for email classification practice. The original data has 10000 instances of 448 features.
    \item \texttt{TIMIT} ($N=390$). It is the covariance matrix of the TIMIT speech recognition data\footnote{available at \url{https://catalog.ldc.upenn.edu/LDC93S1}}. There are 151290 instances, each contains 390 features (concatenating MFCC features in 10 consecutive 30ms windows).
    \item \texttt{Votes} ($N=16$).  It is the covariance matrix of the \emph{Congressional Voting Records} data set in the UCI Repository. The original data has 435 instances of 16 dimensions.
\end{itemize}

\setlength\tabcolsep{1.5pt}

\section{Long Range Arena}
In this section, we provide a detailed explanation of the LRA experiments \citep{tay2020long}. The technical information, such as memory consumption and time per epoch, is provided. In addition, we visualized attention maps for several instances from the Pathfinder task and explained their extraction process in detail.  
\label{appendix:lra}

\setlength\tabcolsep{1pt}
\newcommand{\lraegfigwidthone}{2.6cm}
\begin{figure}[t]
	\begin{center}
	\begin{tabular}{ccc}
		\includegraphics[width=\lraegfigwidthone]{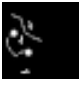} &
		\includegraphics[width=\lraegfigwidthone]{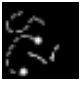} &
		\includegraphics[width=\lraegfigwidthone]{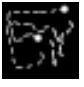}\\
		{\tiny {\bf Easy}} &
		{\tiny {\bf Medium}}&
		{\tiny {\bf Hard}} \\
 	\end{tabular}
	\end{center}
	\caption{Example images form the Pathfinder task.}
	\label{fig:pathfinderex}
	\vspace{-3mm}
\end{figure}

\setlength\tabcolsep{3pt}

\subsection{Data Description}
The data set for the LRA benchmark is publicly available. The information about data and the download link can be found in the official GitHub repository: \url{https://github.com/google-research/long-range-arena}.
\begin{itemize}
    \item \textbf{ListOps} The raw data for this problem is organized as three separate files \texttt{basic\_train.tsv}, \texttt{basic\_test.tsv}, \texttt{basic\_val.tsv} for training, testing, and validation data, respectively. The split is fixed. In addition to the tokens described in the main paper, each sequence has "(" and ")" symbols, which should be removed. To equalize the lengths of the sequences, we used the built-in PyTorch padding functional. After the sequences are prepared, the embedding layer processes each unique value, thus mapping elements to the embedding space. The rest of the training process is straightforward. 
    \item \textbf{Text Classification} The IMDB data set is downloaded using the \texttt{tensorflow-dataset} package. It includes 25\,000 instances for training and another 25\,000 for the test set. To transform the raw reviews into sequences, we first went through the whole corpus and extracted the character vocabulary. Then we mapped each sequence to a vector of indices using this vocabulary. Finally, we truncated or padded each sequence to a fixed length of $4k$ and started training.
    \item \textbf{Image Classification} CIFAR10 is a well-known dataset, which can be downloaded from the \texttt{torchvision} package. The train/test splitting is fixed. To make images gray-scaled, we used standard transformation \texttt{transforms.Grayscale} from the same package. An image is flattened to a sequence of length 1024. Then each element is mapped to a dictionary of size 256 (all possible intensity values) and given to the embedding layer.  
    \item \textbf{Pathfinder} The problem data consists of two types of files: images and metafiles. Metafiles store information about all the images and their corresponding labels (positive or negative). There are three classes of images: \texttt{curv\_baseline} (easy), \texttt{curv\_contour\_length\_9} (medium), \texttt{curv\_contour\_length\_14} (hard). An image class corresponds to the distance between its endpoints (curve length), thus positively correlates with the difficulty level. Three images from the different catalogs are present in Figure \ref{fig:pathfinderex}. The exact data split is not provided. To separate the data into three parts, we iterated over all metafiles from the catalogs and constructed the training/val/test (90\%/5\%/5\%) sets such that all three types of images are present equally. The rest of the processing is similar to the Image Classification task.
\end{itemize}

\setlength\tabcolsep{2.5pt}
\newcommand{\lraegfigwidthtwo}{4cm}
\begin{figure}[t]
	\begin{center}
	\begin{tabular}{ccc}
		\includegraphics[width=\lraegfigwidthtwo]{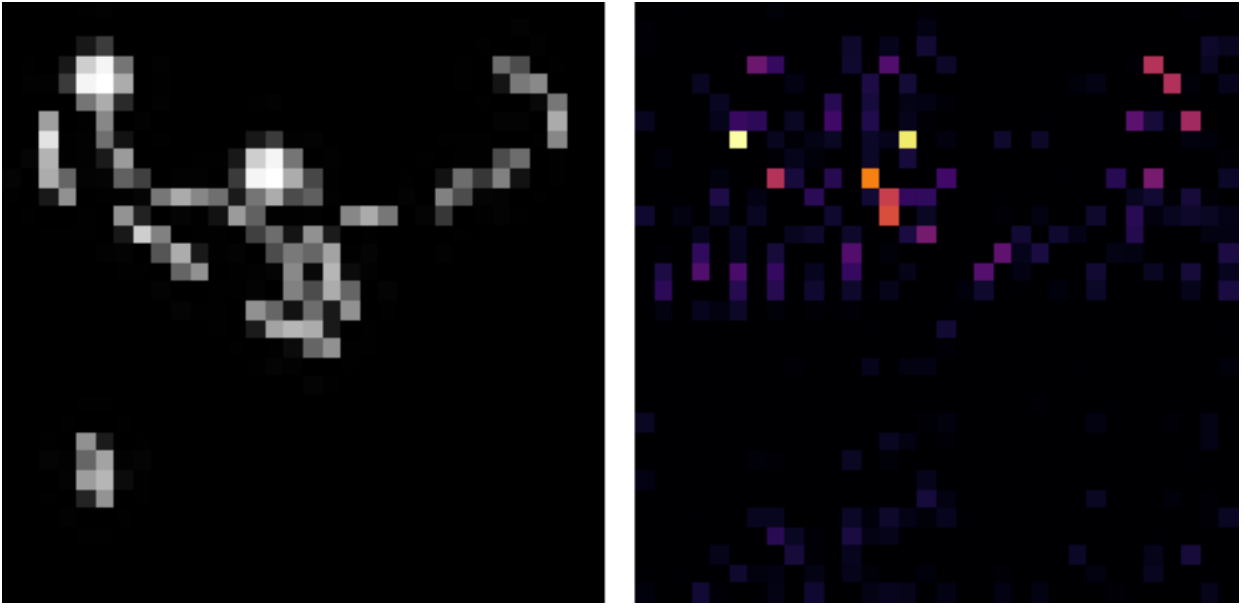} &
		\includegraphics[width=\lraegfigwidthtwo]{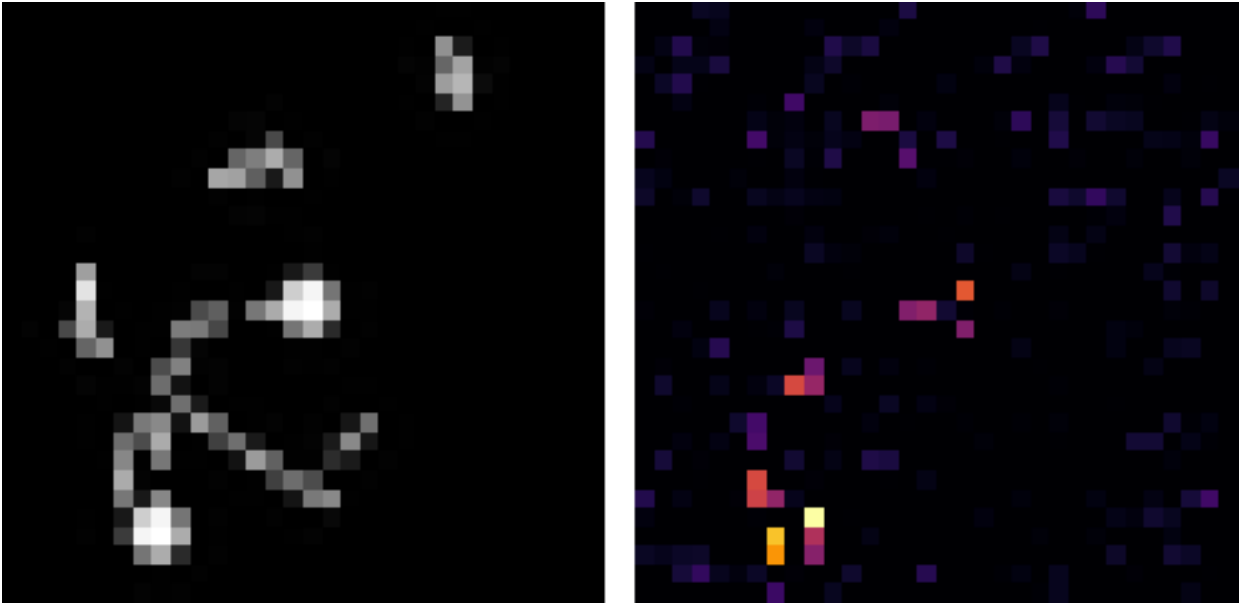} & 
		\includegraphics[width=\lraegfigwidthtwo]{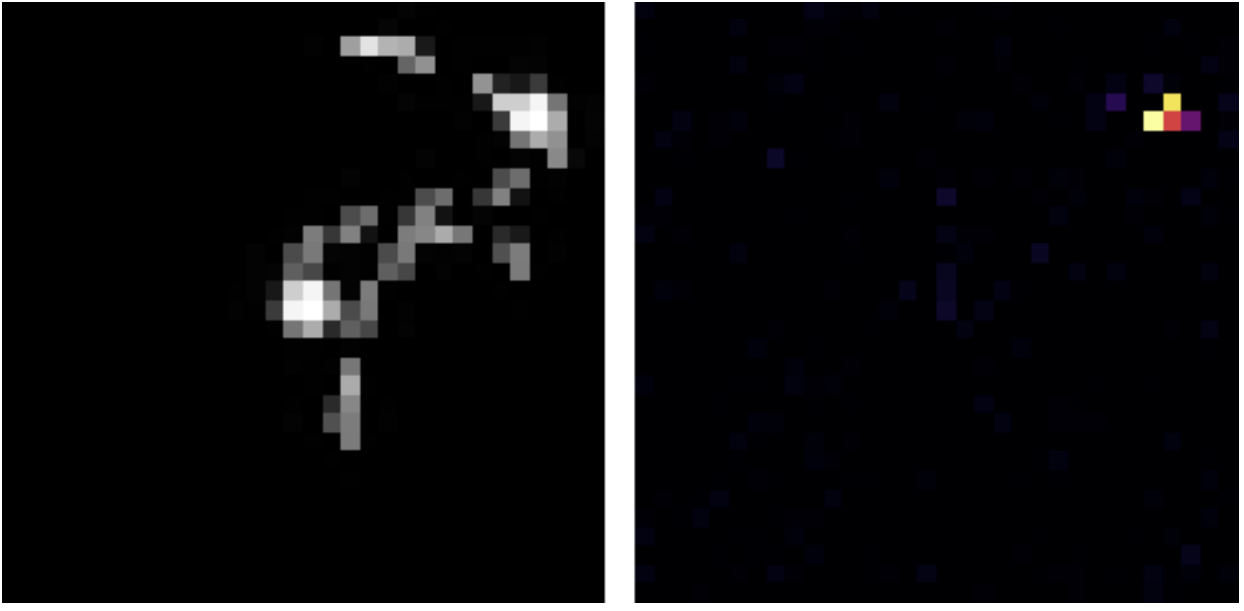} \\
		\includegraphics[width=\lraegfigwidthtwo]{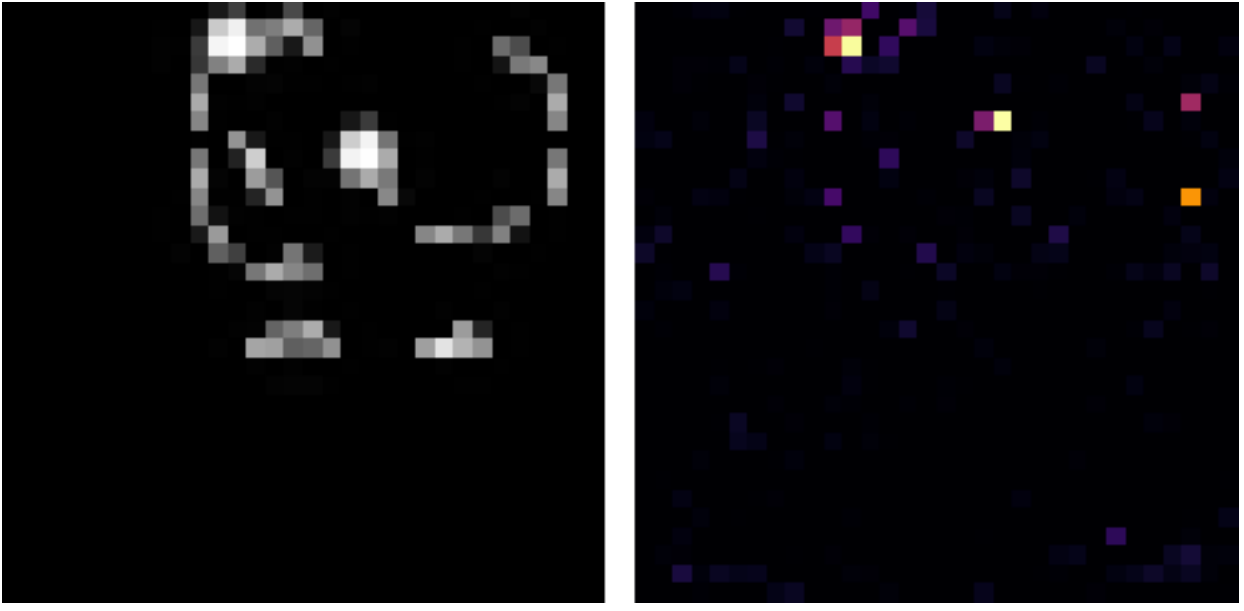} &
		\includegraphics[width=\lraegfigwidthtwo]{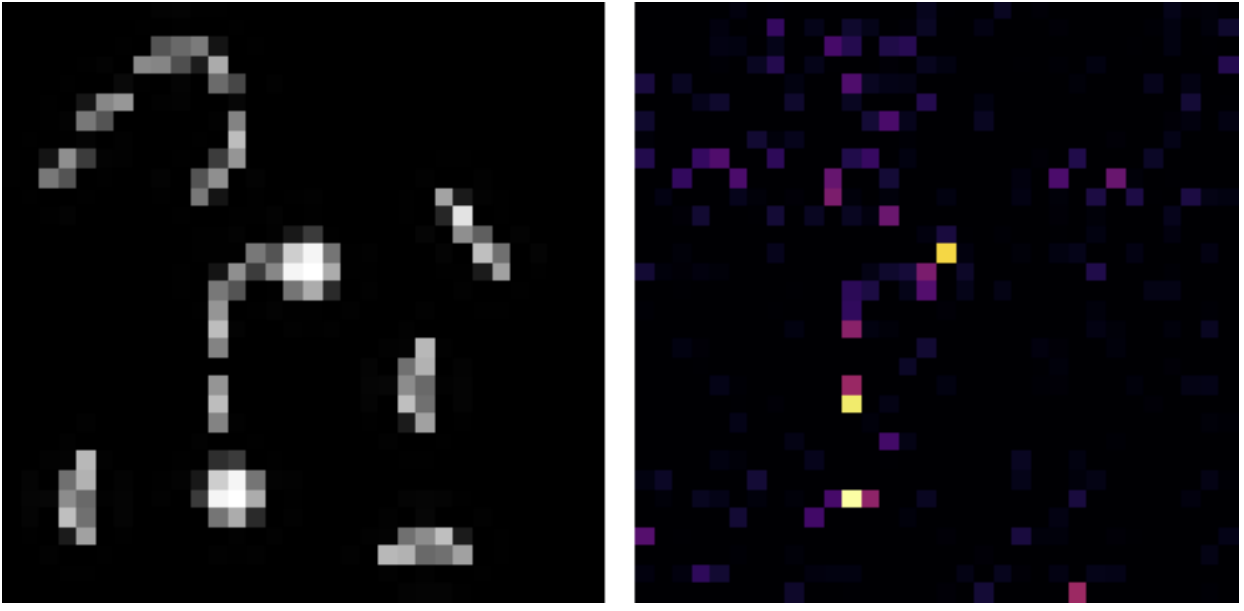} &
		\includegraphics[width=\lraegfigwidthtwo]{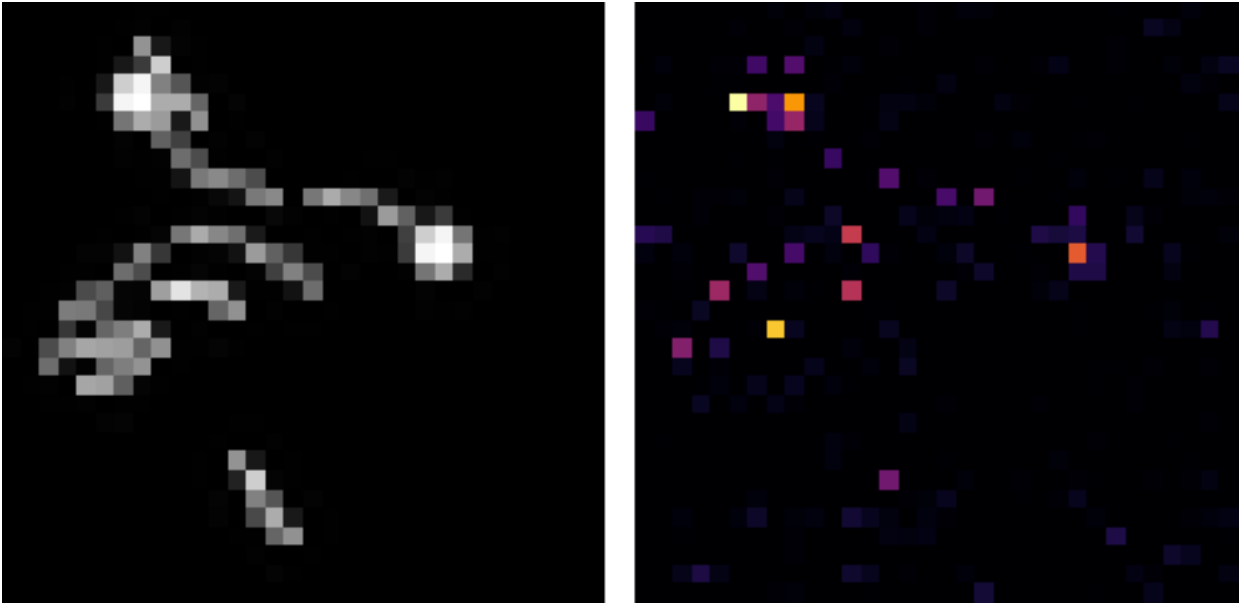}\\
		\includegraphics[width=\lraegfigwidthtwo]{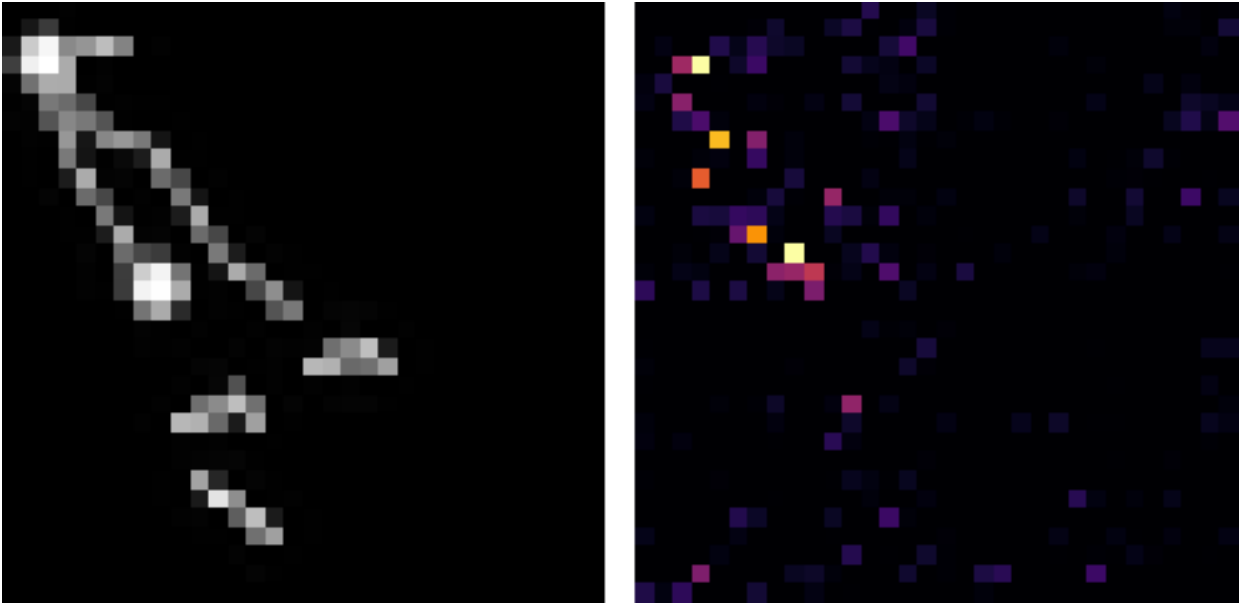} &
		\includegraphics[width=\lraegfigwidthtwo]{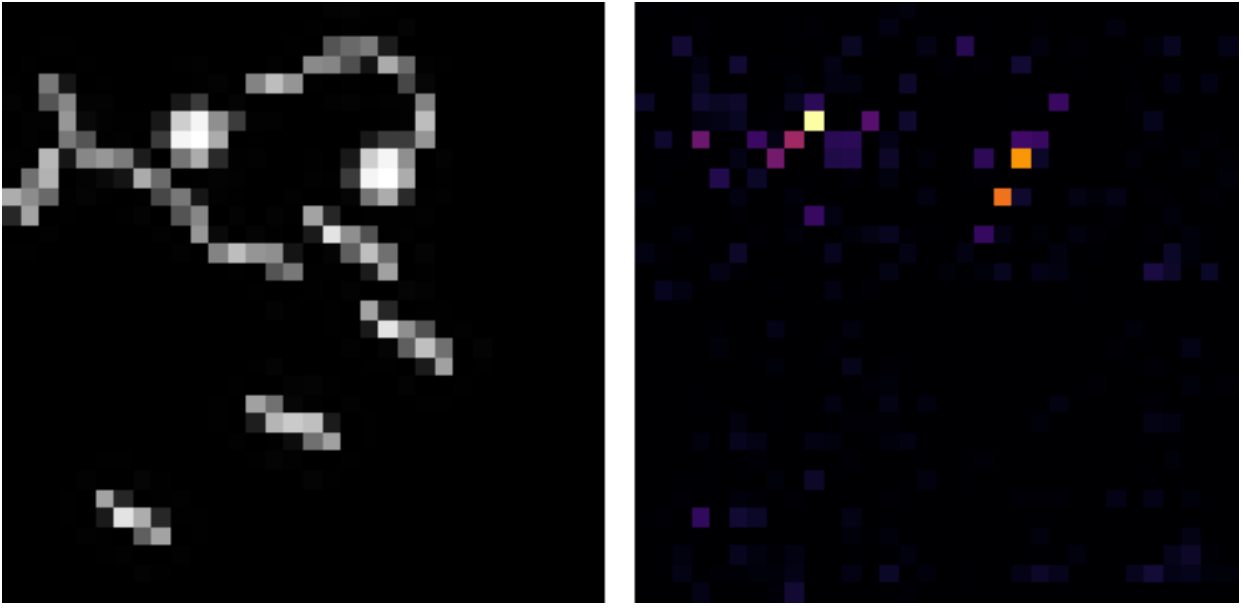} &
		\includegraphics[width=\lraegfigwidthtwo]{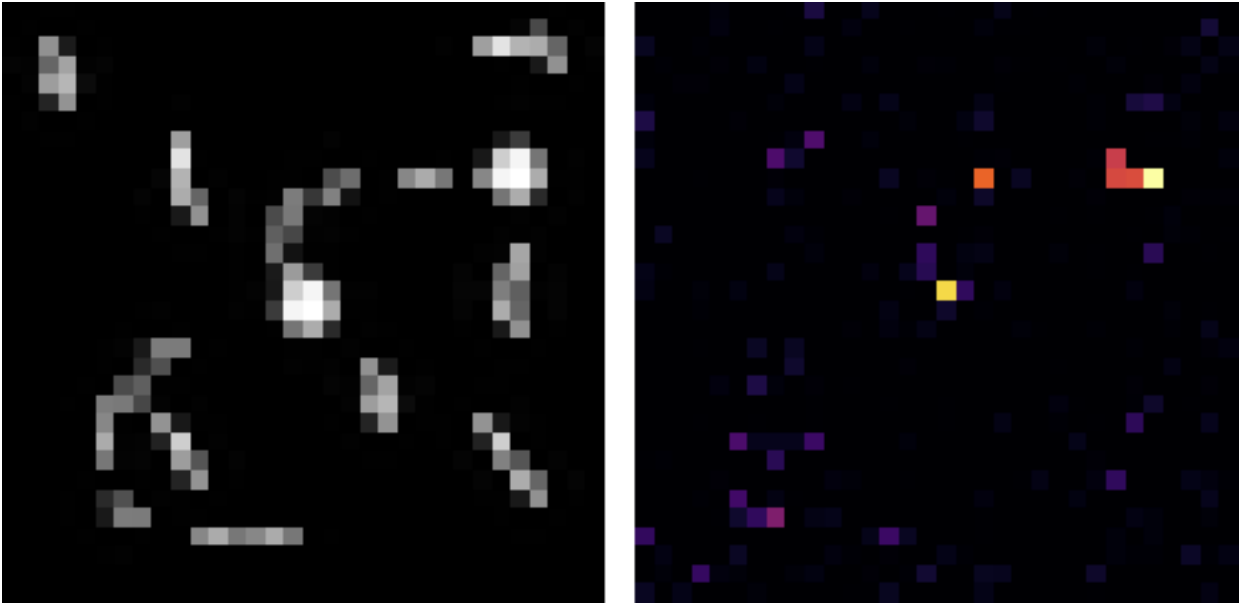} \\
 	\end{tabular}
	\end{center}
	\caption{Example of Pathfinder images and their corresponding attention maps.}
	\label{fig:pathfindermaps}
	\vspace{-3mm}
\end{figure}

\subsection{Visualization of Attention Maps}

In this subsection, we explain the process of attention values extraction in detail. Two different approaches are used to obtain and interpret the training results because PSF-Attn was trained using different pooling strategies: adding the ["CLS"] token to the sequences of the IMDB reviews and simple flattening for the Pathfinder sequences. 

Having an attention matrix $\Xh=\prod_mW^{(m)}$, the attention vector of the $i$-th sequence element is the $i$-th row of $\Xh$.

\begin{itemize}
    \item \textbf{Text Classification} The ["CLS"] token is responsible for representing the information collected from the whole input sequence. Its attention vector corresponds to the $0$-th row in the attention matrix. After normalization of the absolute matrix values to a range of [0, 1], we extracted the row and visualized its values along with the corresponding review characters of one test instance. The example in the main paper illustrates the ability of PSF-Attn to operate on sparse attention matrices successfully and to output meaningful results reflecting high classification accuracy.

    \item \textbf{Pathfinder} 
    Flattening as a pooling strategy requires another approach to interpret the attention values. The attention matrix $\Xh$ for every test instance is of size (1024x1024), while the input image is (32x32). This mismatch makes it problematic to map attention values to the corresponding input image directly. To achieve this, we identified the positions of the endpoints and considered their direct neighbors (18 pixels positions, including themselves) as the source for visualization. We extracted the respective rows of the attention matrix and, finally, took their sum. The resulting vector was reshaped to the original size and depicted in a heat map. Figure \ref{fig:pathfindermaps} shows the attention values obtained using the described method. We can see that the identified attention vectors of $\Xh$ store the meaningful information that visually matches the target path. 
\end{itemize}

\begin{table}[t]
\centering
\caption{Technical details of the PSF-Attn training on all experiments: time per epoch (sec.), number of training parameters, and peak memory usage (GB). Benchmarks are run on a Linux machine with one NVIDIA Tesla V100 32GB, Intel Xeon Gold 6240 CPU @ 2.60GHz processors, with 754GB of system memory.}
\label{tab:timings}
\begin{tabular}{lccccc}
\hline\hline
Problem     & $N$ & Train Size (k) & Time & Param. (M) & Memory\\
\hline
ListOps       & $2000$ & 96 & 431 & 1.86 & 7.90\\
Text          & $4096$ & 25 & 74 & 0.21 & 6.54 \\
Image         & $1024$ & 50 & 21 & 0.29 & 1.73 \\
Pathfinder    & $1024$ & 540 & 374 & 0.17 & 3.58\\
\hline
Adding       & $2^7$    & 200 & 34 & 0.01 & 1.40\\
             & $2^8$    & 200 & 36 & 0.02 & 1.43\\
             & $2^9$    & 200 & 39 & 0.03 & 1.52\\
             & $2^{10}$ & 200 & 45 & 0.06 & 1.73\\
             & $2^{11}$ & 200 & 87 & 0.10 & 2.12\\
             & $2^{12}$ & 200 & 175 & 0.18 & 3.08\\
             & $2^{13}$ & 200 & 377 & 0.35 & 5.12\\
             & $2^{14}$ & 200 & 845 & 0.68 & 9.97\\
\hline
Order        & $2^7$    & 200 & 34 & 0.02 & 1.40\\
             & $2^8$    & 200 & 38 & 0.03 & 1.43\\
             & $2^9$    & 200 & 45 & 0.05 & 1.52\\
             & $2^{10}$ & 200 & 61 & 0.08 & 1.74\\
             & $2^{11}$ & 200 & 104 & 0.15 & 2.14\\
             & $2^{12}$ & 200 & 193 & 0.28 & 3.10\\
             & $2^{13}$ & 200 & 397 & 0.55 & 5.17\\
             & $2^{14}$ & 200 & 871 & 1.07 & 10.47\\
\hline
\hline
\end{tabular}
\end{table}

\end{document}